\documentclass[10pt,twocolumn,letterpaper]{article}

\usepackage{cvpr}
\usepackage{times}
\usepackage{epsfig}
\usepackage{graphicx}
\usepackage{amsmath}
\usepackage{amssymb}

\usepackage{subfigure}
\usepackage{algorithm}
\usepackage{algorithmic}
\usepackage{multirow}
\usepackage{color}
\usepackage{amsmath}
\usepackage{epstopdf}
\usepackage{rotating}

\usepackage{array}
\makeatletter
\newcommand{\thickhline}{%
    \noalign {\ifnum 0=`}\fi \hrule height 0.7pt
    \futurelet \reserved@a \@xhline
}
\newcolumntype{"}{@{\hskip\tabcolsep\vrule width 1pt\hskip\tabcolsep}}
\makeatother

\newcommand{\tabincell}[2]{\begin{tabular}{@{}#1@{}}#2\end{tabular}}

\usepackage{slashbox,pict2e}

\usepackage[pagebackref=true,breaklinks=true,letterpaper=true,colorlinks,bookmarks=false]{hyperref}

\cvprfinalcopy 

\def\cvprPaperID{} 

\ifcvprfinal\fi
\begin{document}

\title{L2GSCI: Local to Global Seam Cutting and Integrating for \\ Accurate Face Contour Extraction}

\author{Yongwei Nie\\
South China University of China\\
{\tt\small nieyongwei@scut.edu.cn}
\and
Xu Cao\\
South China University of China\\
{\tt\small }
\and
Chengjiang Long\\
Kitware, Inc.\\
\and
Ping Li\\
The Education University of Hong Kong\\
\and
Guiqing Li\\
South China University of China\\
}

\maketitle

\begin{abstract}
Current face alignment algorithms can robustly find a set of landmarks along face contour. However, the landmarks are sparse and lack curve details, especially in chin and cheek areas where a lot of concave-convex bending information exists. In this paper, we propose a local to global seam cutting and integrating algorithm (L2GSCI) to extract continuous and accurate face contour. Our method works in three steps with the help of a rough initial curve. First, we sample small and overlapped squares along the initial curve. Second, the seam cutting part of L2GSCI extracts a local seam in each square region. Finally, the seam integrating part of L2GSCI connects all the redundant seams together to form a continuous and complete face curve. Overall, the proposed method is much more straightforward than existing face alignment algorithms, but can achieve pixel-level continuous face curves rather than discrete and sparse landmarks. Moreover, experiments on two face benchmark datasets ({\em i.e.}, LFPW and HELEN) show that our method can precisely reveal concave-convex bending details of face contours, which has significantly improved the performance when compared with the state-of-the-art face alignment approaches.

\end{abstract}

\section{Introduction}

\begin{figure}[t]
\centering
\includegraphics[width=0.49\linewidth]{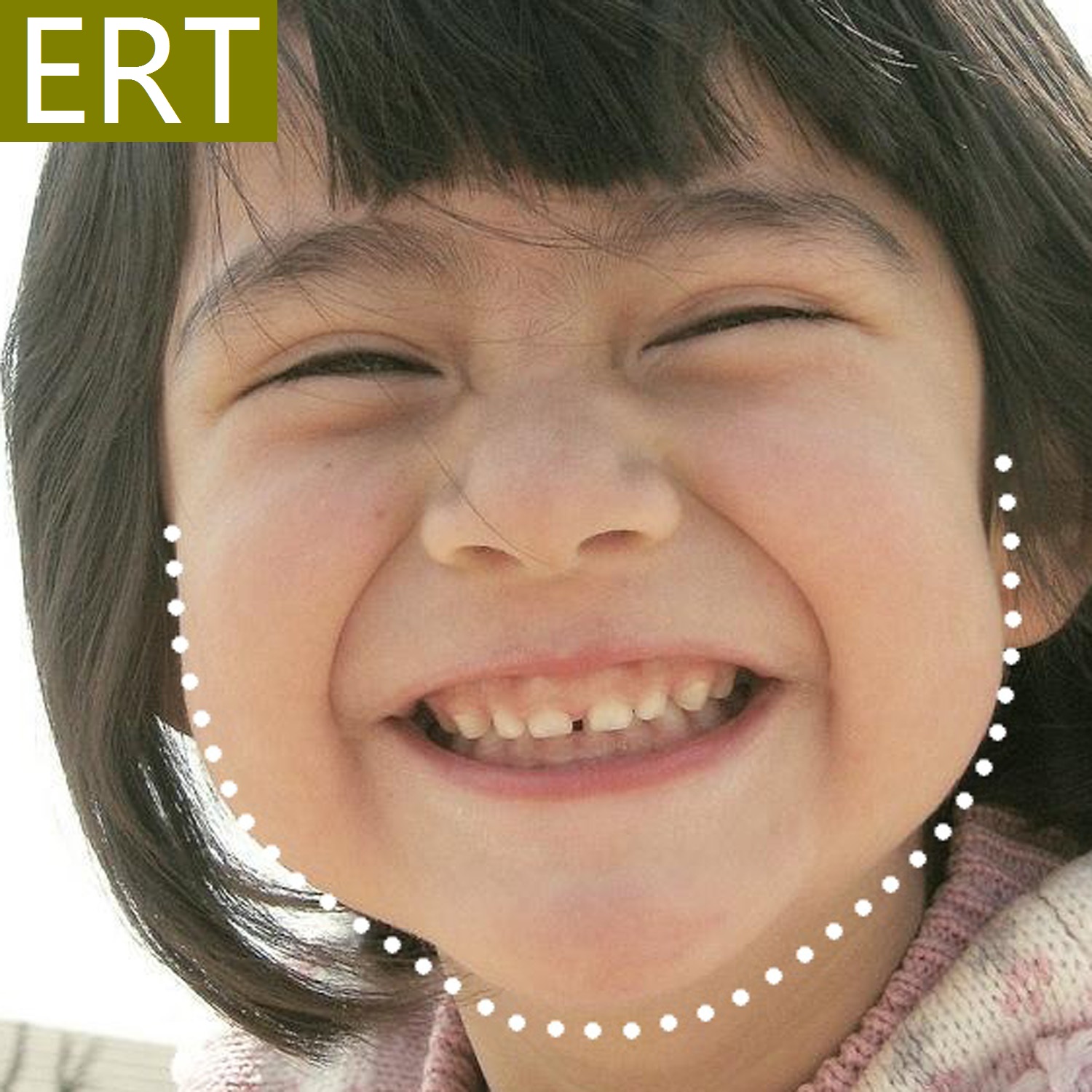}
\includegraphics[width=0.49\linewidth]{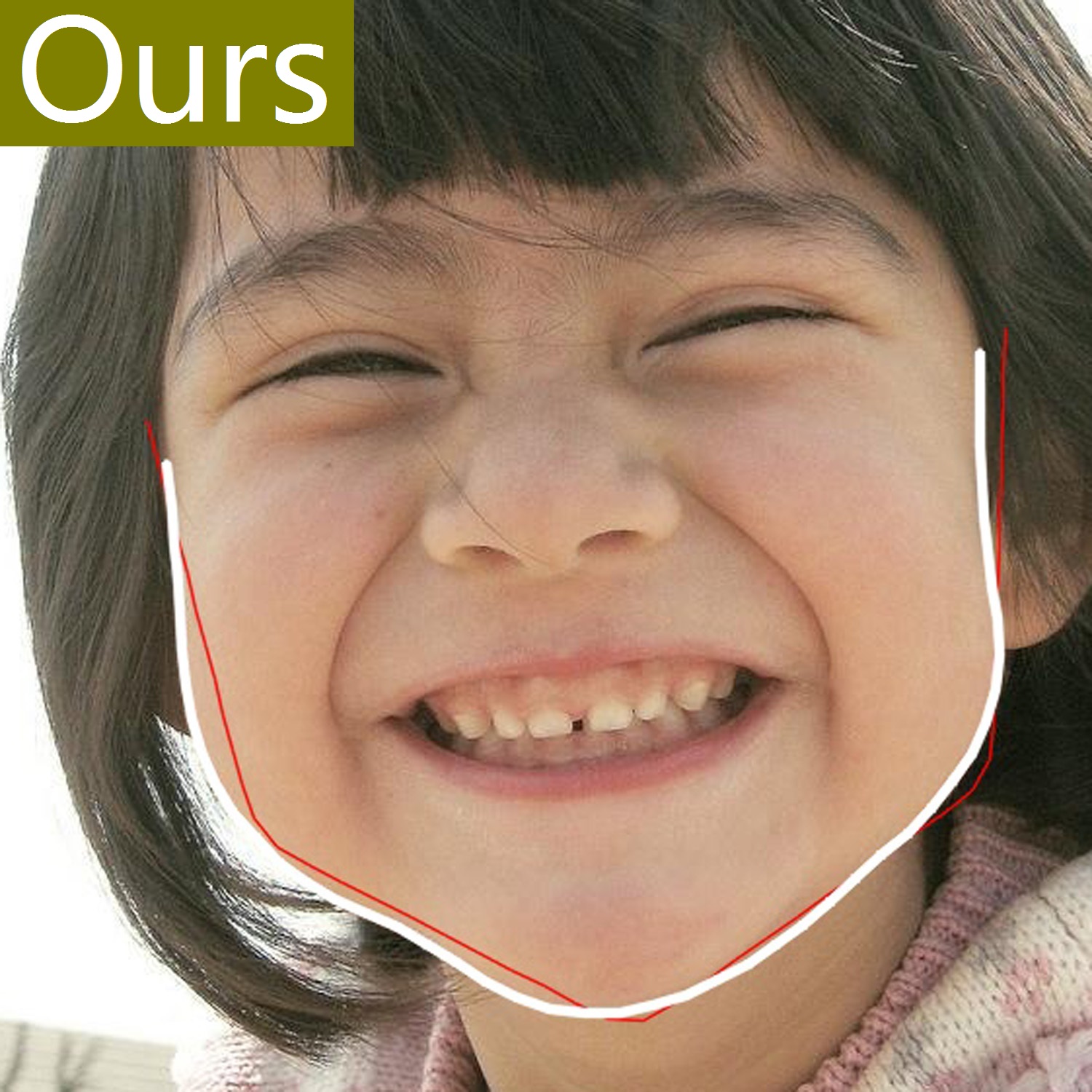}
\caption{From sparse landmarks to continuous face contour. \textbf{Left}: the result of Ensembles of Regression Trees (ERT) \cite{kazemi2014one} which are a set of sparse landmarks. \textbf{Right}: with the help of an initial guess (the red curve), our method directly identify the pixel-continuous curve (white) that tightly fits with the real face contour.}
\label{fig:introd}
\end{figure}

Face alignment tries to identify facial landmarks on eyes, nose, mouth, and face contour. However, it is well recognized that the outer landmarks on the face contour are more variant than their inner counterparts. In this paper, we go a litter farther: rather than a set of discrete and sparse landmarks on face contour, we try to extract the pixel-level continuous face contour itself (see Fig. \ref{fig:introd}).

Face contour in the chin and cheek areas has abundant bending details on how the cheek caves in or the chin bulges. Such details are crucial for the success of many high-level vision tasks, such as face recognition\cite{Masi_2016_CVPR,Sun_2016_CVPR}, emotion representation \cite{Martinez_JMLR2012}, 3D face reconstruction \cite{Liu_ECCV2016}, \etc. Improving the accuracy of extracted face contours is of great meaning and value, which has always been an important research topic in the Computer Vision community. However, it is still very challenging to extract accurate and pixel-level continuous face contours, especially in chin and cheek areas.

Most of recently developed methods focus on the supervised strategies, such as Active Shape Model (ASM) \cite{cootes1995active}, Active Appearance Model (AAM) \cite{cootes2001active}, part-based deformable models \cite{cristinacce2006feature,belhumeur2013localizing,zhu2012face}, and cascaded regression-based models \cite{cootes2012robust,dantone2012real,burgos2013robust,yang2013sieving,cao2014face,ren2014face,kazemi2014one,lee2015face}. However, although supervised algorithms have greatly improved the robustness of facial landmark identification, the obtained landmarks are very sparse, and therefore are difficult to describe the concave-convex details of chin and cheeks flexibly.

\begin{figure*}[t]
    \begin{center}
    \includegraphics[width=1\linewidth]{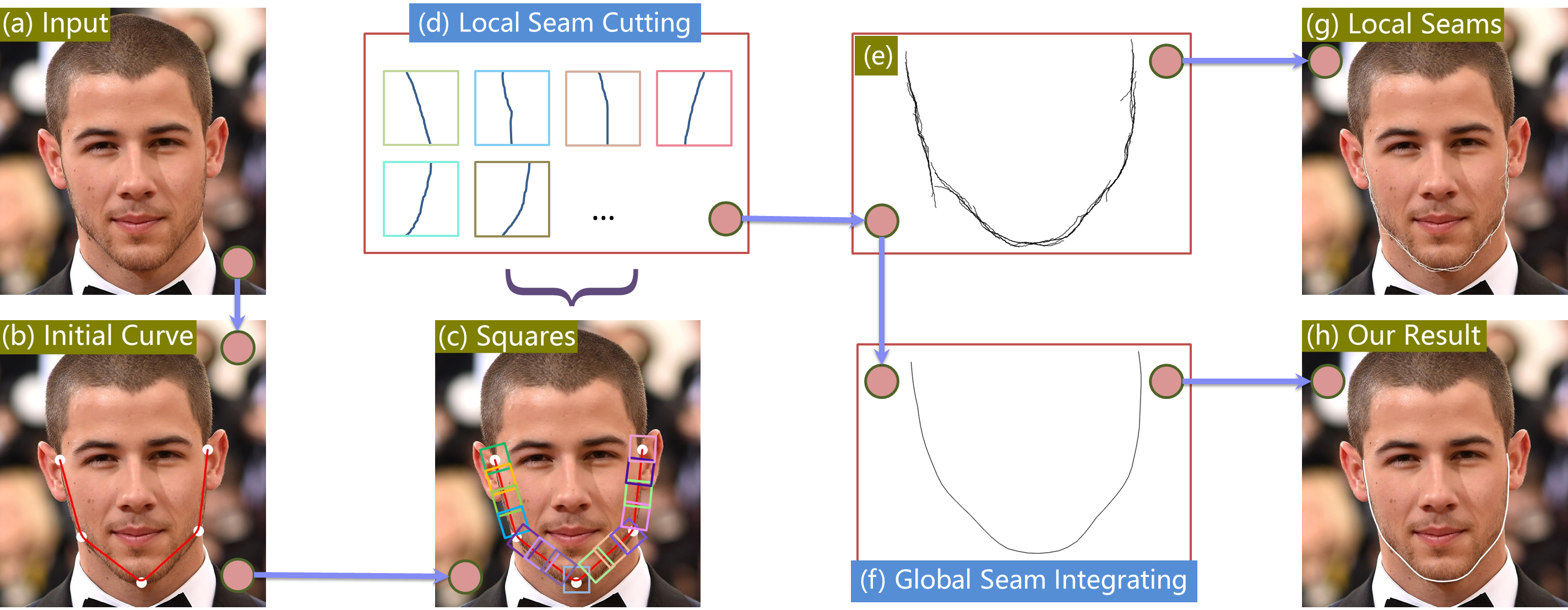}
    \end{center}
    \caption{Overview of the proposed algorithm. (a) The input face. (b) Initial curve produced by ERT \cite{kazemi2014one} which is trained with only 5 landmarks, \ie, two landmarks beside eyes, two beside mouth, and one near the chin. (c) Squares sampled along the initial curve. (d) The proposed gradient and parabola guided local seam cutting method can extract the local segment of the target face contour in each square region. (e) All the local seams. (f) The proposed PCA-based global seam integrating method can connect all local seams robustly to obtain a complete face curve. (g) Local seams overlaid on the input image. (h) Our extracted face contour overlaid on the input image.}
    \label{fig:overview}
\end{figure*}

In this paper, we propose a local to global seam cutting and integrating algorithm (L2GSCI) to directly find continuous face curves, and to reveal concave-convex bending details in chin and cheek areas. We take a very different design strategy from those of the existing object contour detection \cite{kass1988snakes,bertasius2015deepedge} or face alignment approaches \cite{cootes2001active,zhu2012face,kazemi2014one}. We find that most of the existing approaches smoothly refine an initial guess to approach the target contour in an iterative way. However, such a process may be easily impeded by any obstacle in the halfway. We want to avoid this problem by directly finding the target face contour. To achieve the goal, our proposed method is mainly based on the observation that each small segment of a face contour usually has large gradients and can be well fitted by a parabolic curve. We take advantage of this characteristic, and divide the face contour into local segments. We then propose a gradient and parabola guided seam cutting method to directly identify each local segment, and finally integrate all the local segments to obtain a global and complete face contour by a robust and non-parametric medial abstraction method based on principal components analysis (PCA). Our method is facilitated by the recent developments in the face alignment literature, as they can provide us with a nearly correct initial face curve (the red curve in Fig. \ref{fig:introd} (b)) which helps locating and segmenting local segments of the target face contour.

To summarize, our contributions in this paper are:
\vspace{-0.08in}
\begin{itemize}
\item We propose a three-step local to global seam cutting and integrating algorithm (L2GSCI) that is able to extract precise face contours based on the parabola assumption of local segments of face contours.
\vspace{-0.08in}
\item Our method explores pixel-level continuous curves which are more informative and flexible to reveal concave-convex bending details of the ground truth face contours than a set of discrete landmarks.
\vspace{-0.08in}
\item We demonstrate experimentally that our method provides a very economical way to significantly improve the results of the state-of-the-art face alignment approaches.
\end{itemize}

\section{Related Work}
\label{sec:related}
Related prior work can be roughly divided into two categories: object contour detection and face alignment. We now review the related work and clarify the key difference from our approach.

{\bf {Object contour detection}} is a common task for image interpretation. In the early stages, researchers employed deformable templates or flexible models \cite{kass1988snakes,beinglass1991articulated,yuille1992feature} to fit salient image contours. For example, Kass \etal \cite{kass1988snakes} presented the Active Contour Models (ACM) or snakes which can approach target contour by minimizing an energy function. Recently, researchers extracted object contours by techniques of clustering, segmentation, or deep-learning \cite{arbelaez2011contour,bertasius2015deepedge}. The difference between those methods and ours is that they usually try to extract more general object contours, or say edges, while our method is designed to extract contours of faces. A few methods \cite{perlibakas2003automatical, bing2004face} extended the internal and external energies of ACM \cite{kass1988snakes} to process face contour. However, their experiments were taken on very simple examples, and cannot achieve satisfactory results on modern face alignment datasets.

{\bf {Face alignment}} may date back to the ASM \cite{cootes1995active} and AAM \cite{cootes2001active} proposed by Cootes \etal, in which a joint point distribution model is built as a PCA based linear regression problem to constrain the shape variations of face landmarks. Then, many follow-up works, such as boosted regression \cite{saragih2007nonlinear} or direction classifiers \cite{tresadern2010additive}, extended AAM from linear to non-linear to improve accuracy. Later, instead of using just a global shape constraint like in AAM-based methods, part-based deformable models were proposed to optimize an objective function comprised of both a shape prior term and local experts terms \cite{cristinacce2006feature,belhumeur2013localizing,zhu2012face}. Method in \cite{alabort2015unifying} unified AAM-like holistic models \cite{papandreou2008adaptive} and part-based models \cite{saragih2011deformable} together, and obtained good results.

Nowadays, the mainstreams in face alignment are based on regression which directly model the mapping from local features to face shapes \cite{dollar2010cascaded,cootes2012robust,dantone2012real,burgos2013robust,yang2013sieving,cao2014face,ren2014face,kazemi2014one,lee2015face}. We can discriminate these methods by what kind of regressors or features they employed. The above face alignment algorithms have been well established to find facial landmarks. However, the found landmarks are sparse, and are not flexible enough to reveal more detailed concave-convex bending information that is necessary to generate accurate face contours. Recently, more efforts on face alignment algorithms were taken to handle very challenging cases, such as very large poses or severe occlusions, by deep learning \cite{zhang2016learning,Jourabloo_2016_CVPR,Zhang_2016_CVPR}, additional 3D information \cite{Zhu_2016_CVPR}, or leveraging multiple datasets \cite{zhang2015leveraging}. Different from them, our aim is to further improve the extraction accuracy of the visible part of face contour.

\section{Algorithm}
\label{sec:algorithm}
As shown in Fig. \ref{fig:overview}, the whole pipeline of L2GSCI consists of the following three steps: (1) we sample overlapped squares along the initial guess curve to divide the target face contour into small segments; (2) then we develop a gradient and parabola guided seam cutting method to identify each segment in a local square region; and (3) we finally integrate all the local segments to obtain the final non-parametric face contour by medial curve abstraction. We describe each step in detail in the following subsections.

\subsection{Square Sampling}
\label{sec:square-sampling}
Our algorithm assumes a rough initial guess of the target face contour. In practice, an initial guess can be obtained by connecting face landmarks computed by any regression-based face alignment algorithm \cite{cao2014face,kazemi2014one,ren2014face}. In this paper, we employ the method of Ensembles of Regression Trees (ERT) \cite{kazemi2014one}, and train ERT with only 5 landmarks, \ie, two landmarks beside eyes, two beside mouth, and one near the chin. Fig. \ref{fig:introd} and Fig. \ref{fig:overview} (b) show two examples of the initialization.

Given the initial curve, we divide the target face contour into small and overlapped segments. Our method samples small and overlapped squares along the initial curve. The centers of the squares are on the initial curve, and the square size is set as $0.2\times $ the size of face bounding box in default. As the initial guess is not too far from the real contour, the squares together can contain the target face contour in themselves (as shown in Fig. \ref{fig:overview} (c)). We densely sample face contour segments (70 squares in default) to let them have a lot of overlap, which brings redundancies that benefit for outlier removing in the global step.

As shown in Fig. \ref{fig:overview} (c), we rotate each square around its center to let it align with the tangent direction of the initial curve. This action is useful for the upcoming local segment extraction.

\subsection{Gradient and Parabola Guided Local Seam Cutting}
Now we have divided a face contour into a lot of overlapped squares, and each square contains a segment of the real contour. In this section, we describe how to directly identify the segment within each square region.

As indicated in the previous subsection, each square was rotated according to the corresponding local tangent direction of the initial guess curve. Now, we crop the square of image, and rotate it back to be axis-aligned. This treatment guarantees that the segment in each square is a continuous curve from the top side of the square to the bottom side (see Fig. \ref{fig:overview} (d)).

\begin{figure}[t]
\begin{center}
    \includegraphics[width=0.45\linewidth]{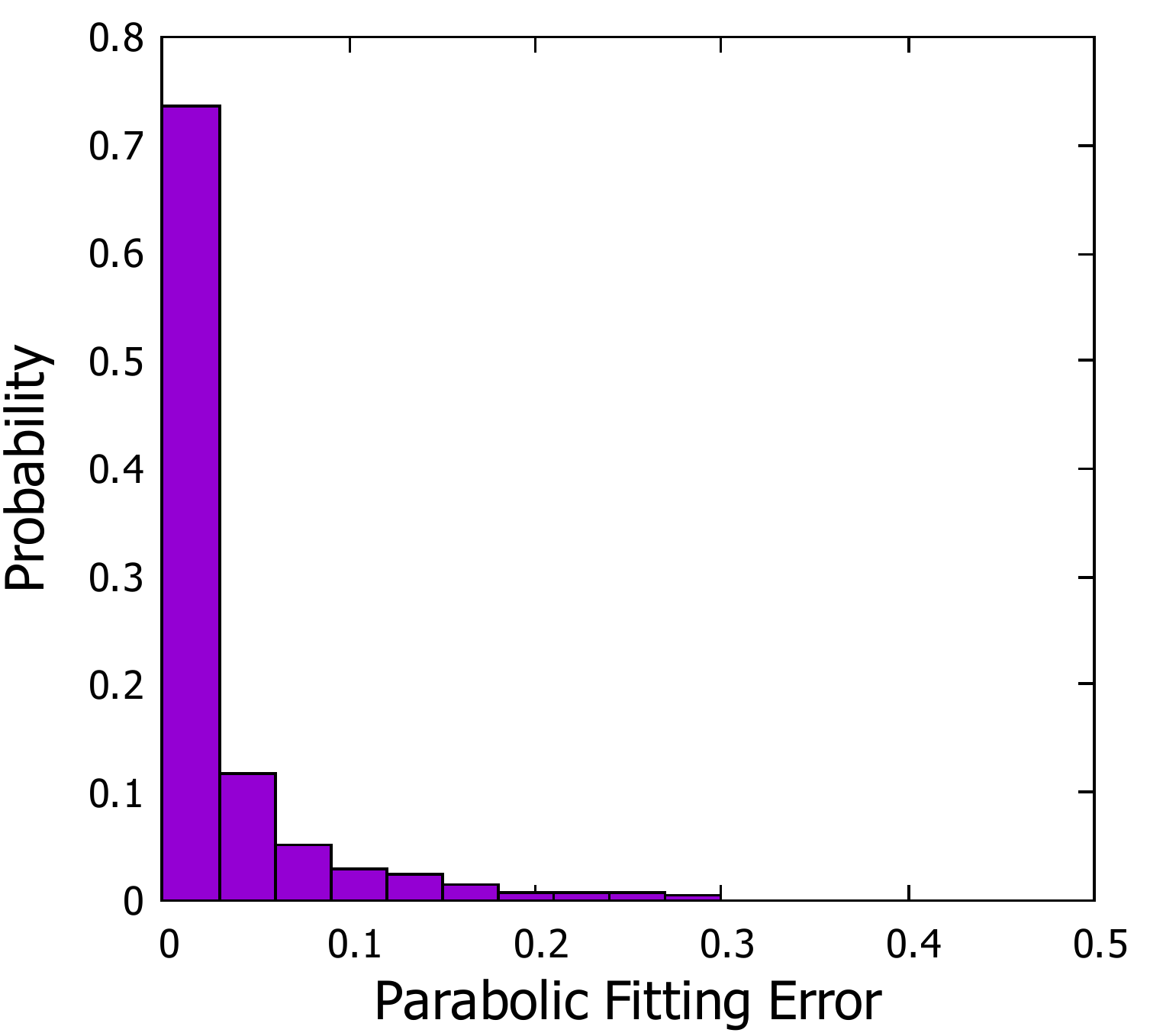}
    \includegraphics[width=0.45\linewidth]{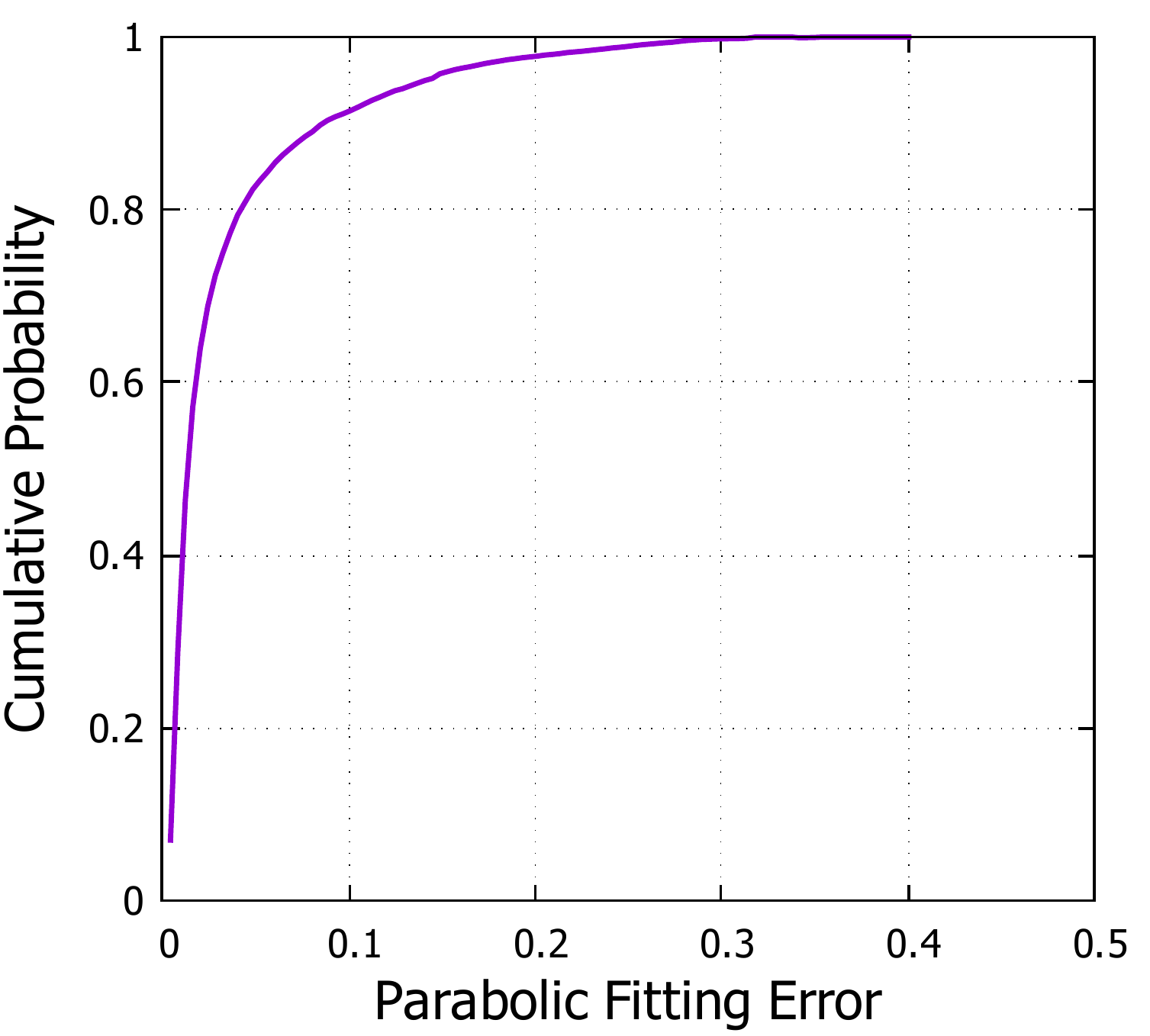}
\end{center}
    \caption{We have 20,000 local segments in total sampled from face contours, together with their fitting errors compared with the fitted parabolic curves. {\bf Left:} the probability of fitting errors. {\bf Right:} the cumulative probability of fitting errors.}
    \label{fig:statis}
\end{figure}

To extract local segments, we take advantage of two characteristics of face contours. Firstly and apparently, a face contour usually has large gradients. Secondly, a statistical analysis points out that local segments of face contours can usually be reproduced by quadratic curves. Our statistical study is conducted as follows. We randomly select 400 face images from two popular face alignment datasets LFPW \cite{belhumeur2013localizing} and HELEN \cite{le2012interactive}. We then annotate ground truth face contours for these images manually, and sample squares to obtain 20,000 local segments. We fit the segments with parabolic curves, and compute the fitting error for each segment. Fig. \ref{fig:statis} shows that about 80\% segments are within the fitting error of 0.05, which demonstrates that these local segments can be well fitted by parabolic curves.

Based on the above observations, we view the local segment to be extracted as a continuous seam $C$ from the top to bottom side that has the most prominent gradients while at the same time is smoothest as a parabola. Formally, let a square region be a $N\times N$ pixel matrix. A possible solution of $C$ is made up of $N$ pixels each of which comes from a row of the pixel matrix:
\begin{equation}
\label{eq:segment}
C=<p_1, p_2, ..., p_i, ..., p_N>,
\end{equation}
where $p_i=(i,j)$ is the coordinate of the $i^{th}$ point of curve $C$, $i$ is row index, and $j\in[1,N]$ is column index. For any pixel $p_i$ and $p_{i+1}$ we require that the difference between their column indices is not greater than 1 to make adjacent points of $C$ to be continuous in pixel grid space.

Let us first only consider the fact of gradient. The seam with maximum gradient energy can be identified by optimizing the following objective function:
\begin{equation}
C^* = \arg \mathop {\max }\limits_{{C}} G({C}) = \arg \mathop {\max }\limits_{{C}} \sum\limits_i {g({p_{i}})},
\label{eq:maxContour}
\end{equation}
where $g(p_{i})$ is the gradient magnitude of $p_{i}$, and $G(C)$ is the total gradient energy of the curve $C$. The above objective function can be optimized by a simple dynamic programming strategy, {\em i.e.},
\begin{equation}
\begin{split}
M(i,j) =
\left\{
\begin{array}{*{20}{l}}
    {g(1,j)}                                         &{i = 1}\\
    {\mathop {\max }\limits_{\Delta}\left\{g(i,j) + M(i - 1,j + \Delta) \right\} }&{i > 1}
\end{array}
\right.
\end{split}
\label{eq:Mij}
\end{equation}
where $\Delta \in \{-1,0,1\}$, and $(i,j)$ is pixel coordinate. There are many curves ending at $(i,j)$. Let $C(i,j)$ be the one among them that has maximum gradient energy. With the help of matrix $M$ which contains such maximum gradient energy at entry $(i,j)$, we can obtain $C(i,j)$ by backtracking from $(i,j)$. Let ${j^*} = \arg \mathop {\max }\limits_j M(N,j)$. The seam $C^*$ from the top side to bottom side with the maximum gradient energy is $C(N,j^*)$.

\begin{figure}[t]
\centering
\includegraphics[width=.24\linewidth]{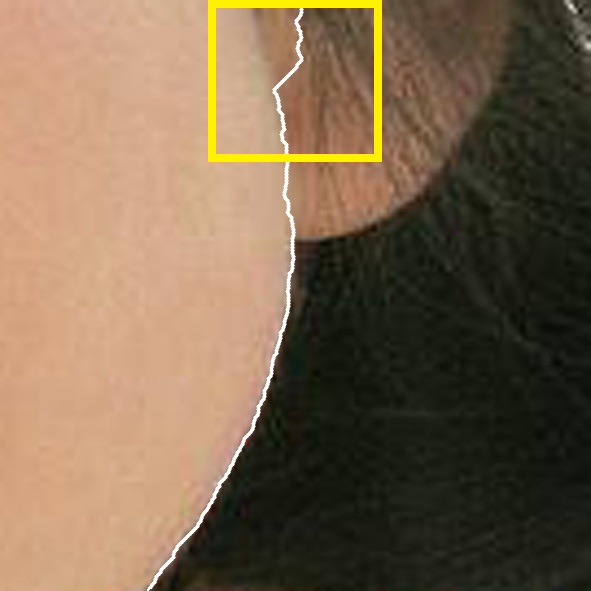}
\includegraphics[width=.24\linewidth]{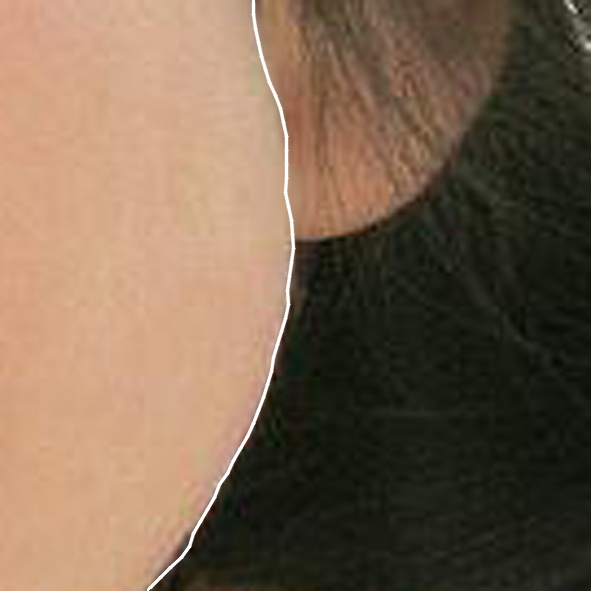}
\includegraphics[width=.24\linewidth]{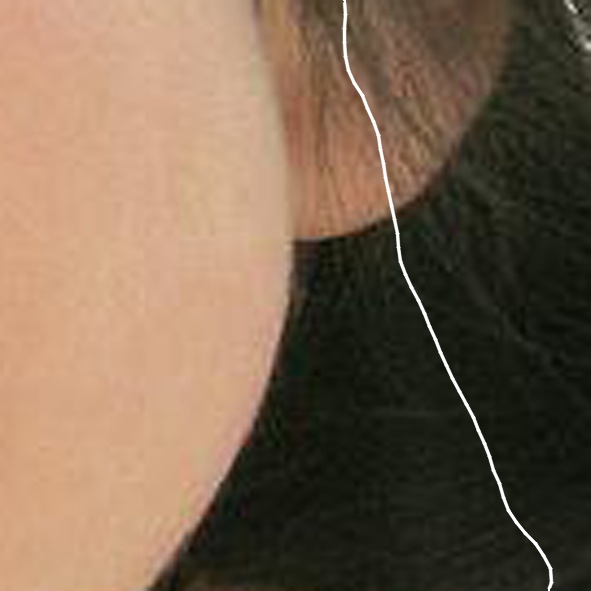}
\includegraphics[width=.24\linewidth]{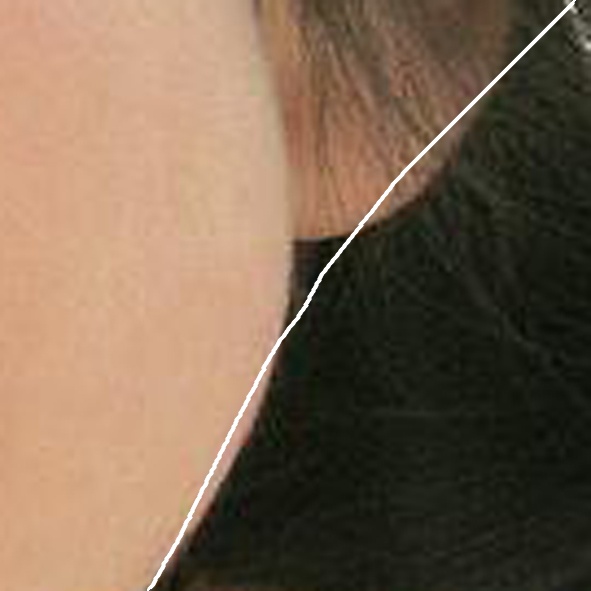}
(a) $\alpha=1.0$ \hspace{0.25cm} (b) $\alpha=0.7$ \hspace{0.25cm} (c) $\alpha=0.4$ \hspace{0.25cm} (d) $\alpha=0.1$
\caption{Local seam cutting results given different values for $\alpha$. (a) Consider gradient only by setting $\alpha=1.0$. Some places (\eg, in the yellow box) may be pulled away toward other areas with larger gradients. (b) A good balance is obtained by setting $\alpha=0.7$. (c) The balance is biased to smoothness by setting $\alpha=0.4$, and the extracted seam would not follow up with maximum gradient. (d) Without consideration of gradient by setting $\alpha=0.1$, the seam becomes a pure parabola.}
\label{fig:alpha}
\end{figure}

However, as shown in Fig. \ref{fig:alpha} (a), the seams extracted by only considering gradient is squiggly and deviates from the real contour, as the gradients at some parts of the real contour are weak, and the dynamic programming progress may be easily pulled into a wrong direction by other areas with larger gradients. To avoid the problem, we utilize the prior that local segments can be well fitted by parabolic curves, and improve the above algorithm by taking better balance between gradient and curve smoothness. Formally, our gradient and parabola guided algorithm optimizes the following energy function that finds the best local seam which has maximum gradient while at the same time is as smooth as possible like a parabola:
\begin{equation}
{\hat{C}} = \arg \mathop {\max }\limits_C \sum\limits_i {(\alpha \cdot g({p_i}) + (1-\alpha) \cdot e({p_i}))}
\label{eq:computeC}
\end{equation}
where $e({p_i})$ is local parabolic fitting error, and $\alpha$ is the weight to adjust smoothness of the computed segment. The smaller the $\alpha$, the smoother the segment. Here, we define the local parabolic fitting error only according to points before $p_i$. Formally, we get a set of 20 points before $p_i$, \ie, $\{p_{i-20},p_{i-19},...,p_{i-1}\}$, and fit a parabolic curve from the set of points. We then compute the nearest distance $d_i$ from $p_i$ to the fitted parabola. Finally, we define $e({p_i}) = 1-{(\frac{d_i}{3})^2}$.

To solve Eq. \ref{eq:computeC} by dynamic programming, we modify the matrix $M$ to be,
\begin{equation}
\label{eq:MijNew}
\begin{split}
M(i,j) =
\left\{
\begin{array}{*{20}{l}}
{g(1,j)}                                                                    &{i = 1}    \\
{\mathop {\max }\limits_{\Delta}\left\{
    \begin{array}{*{20}{l}}
        g(i,j) \\
        + M(i - 1,j + \Delta)
        \end{array}
        \right\} }&{1<i\le 20} \\
{\mathop {\max }\limits_{\Delta}\left\{
    \begin{array}{*{20}{l}}
        g(i,j) \\
        + e_{i-1,j+\Delta}(i,j) \\
        + M(i - 1,j + \Delta)
    \end{array}
    \right\} }&{i>20}
\end{array}
\right.
\end{split}
\end{equation}
where $e_{i-1,j+\Delta}(i,j)$ is the parabolic fitting error of point $(i,j)$ relative to seam $C(i-1,j+\Delta)$.

Note that the parameter $\alpha$ controls the shape of the identified local segments. Fig. \ref{fig:alpha} (a)-(d) show different results with different values of parameter $\alpha$. Experimental results show that $\alpha = 0.7$ is a good balance between gradient and smoothness. Fig. \ref{fig:overview} (e) and (g) show all the extracted local segments of the input face image.

\subsection{PCA-based Global Seam Integrating}

Recall that we have divided the target face contour into segments by densely sampling squares (let $M$ be the number of the squares), and have directly identified each segment in each small square. Let $p^k_i$ be the $i^{th}$ point of the $k^{th}$ segment, we have extracted a total of $M\times N$ points $P = \{p^k_i\}_{i\in [1,N], k\in [1,M]}$. Thanks to the choice of overlapped sampling, the set of points $P$ is highly redundant especially in overlapped regions, and contain both inliers and outliers with respect to the ground truth face contour. Our goal is to find a global curve from $P$ with only one-pixel width while rejecting outliers as much as possible.

To extract the continuous curve, our algorithm should satisfy two requirements. First, the curve should be represented in a non-parametric way. We argue that a face contour usually has abundant but subtle concave-convex details, which are difficult to be described by a simple parametric curve (\eg, spline). Therefore, we choose to directly use a sequence of points $Q=\{q_l\}_{l\in [1,L]}$ to represent our extracted face contour in a non-parametric way, and require adjacent points $q_l$ and $q_{l+1}$ to be as close as possible in Euclidean space, where $L$ is the length of our contour. Second, we strictly require each point $q_l$ to directly come from the point set $P$, as the points in $P$ are carefully identified in the last subsection and have greater gradients, which are more likely on the real face contour. The above two requirements rule out most of the curve skeleton extraction methods in the geometric processing literature \cite{huang2013l1}.

Instead, we design an algorithm which satisfies the above requirements and is robust against outliers. Our method is based on classical weighted PCA. For each point $p^k_i$ of $P$, we construct the weighted covariance matrix of $p^k_i$ with respect to all the other points of $P$:
\begin{equation}
\begin{split}
{Cov}^k_i =
\sum\limits_{i',k'}{\theta (||p^k_i - p^{k'}_{i'}||){{(p^k_i - p^{k'}_{i'})}^T}(p^k_i - p^{k'}_{i'})}
\end{split}
\label{eq:cov}
\end{equation}
where, $i'\in [1,N] \backslash \{i\}$, $k'\in [1,M]\backslash \{k\}$, and $\theta (r)=e^{-r^2/{(h/2)}^2}$ is a fast decaying smooth function with support radius $h = 20$ which is used to reduce the weights of farther points. We then compute the eigenvalues $\{\lambda _0,\lambda _1\}$ of ${Cov}^k_i$ with $\lambda_0 \le \lambda _1$. Finally, we define a measure called directionality degree $\sigma^k_i$ of point $p^k_i$:
\begin{equation}
{\sigma ^k_i} = \sigma (p^k_i) = \frac{{\lambda _1}}{{\lambda _0 + \lambda _1}}
\label{eq:yita}
\end{equation}
The closer $\sigma ^k_i$ is to 1, the smaller $ \lambda_0$ is compared to $ \lambda_1$, and hence the more points around $p^k_i$ are aligned consistently without outliers \cite{huang2013l1}.

\begin{figure}[t]
\centering
\includegraphics[width=0.45\linewidth]{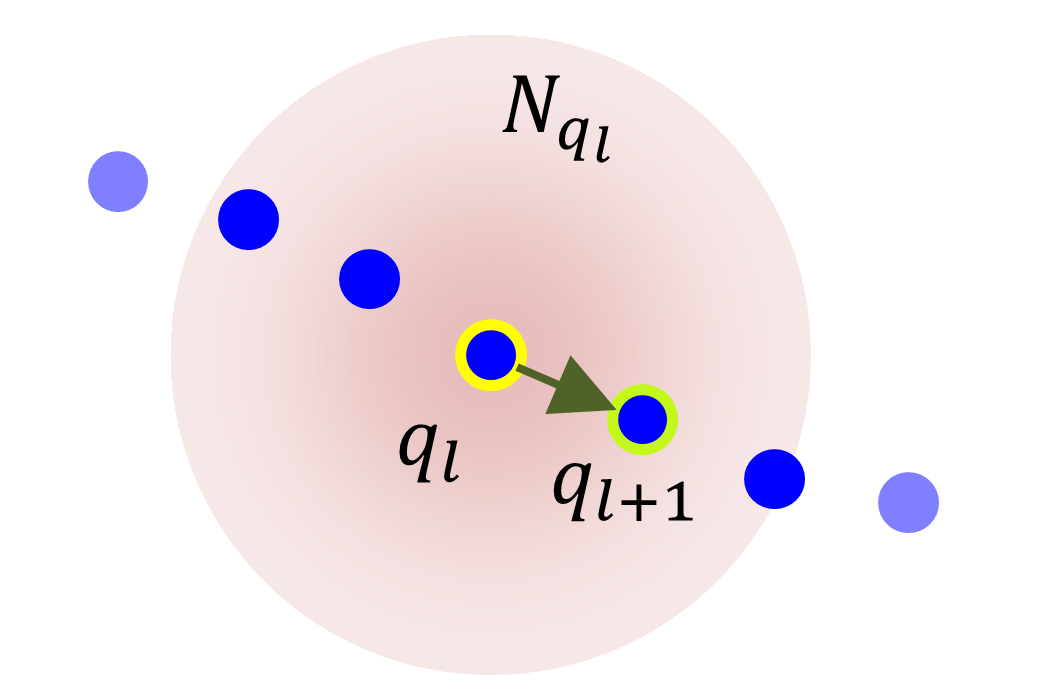}
\includegraphics[width=0.45\linewidth]{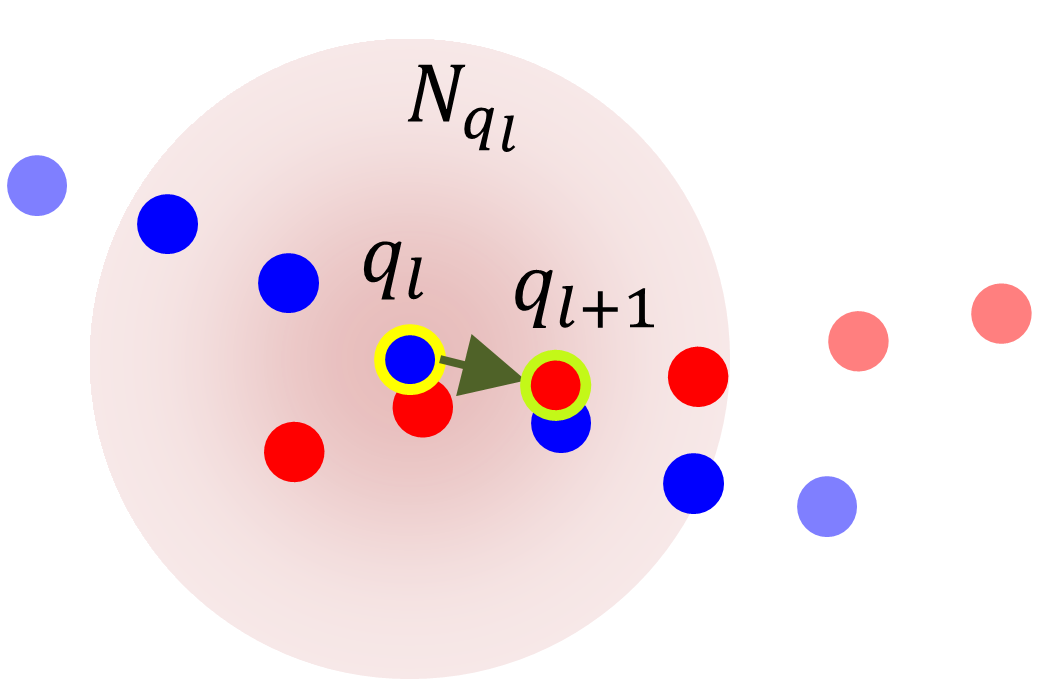}
\caption{\textbf{Left:} When all points in $N_{q_l}$ come from the same segment (all are blue), $q_{l+1}$ is the next point of the segment. \textbf{Right:} Otherwise, we find the one that best matches with the directionality degree of $q_l$.}
\label{fig:pca}
\end{figure}

Given the directionality degree of each point $p^k_i$, our algorithm works as follows. The input to our method is the set of points $P = \{p^k_i\}_{i\in [1,N], k\in [1,M]}$, and the corresponding tangential vectors $V = \{v^k_i\}_{i\in [1,N], k\in [1,M]}$ with respect to locally extracted seams. The output is a face contour $Q = {\{ {q_l}\} _{l \in [1,L]}}$. The first point $q_0$ of $Q$ equals to $p^0_0$. To find $q_{l+1}$ given $q_l=p^k_i$, we first find K-nearest ($K=7$ in default) neighborhood $N_{q_l}=\{p^{k'}_{i'}\}$ of $q_l$. Then two cases are considered. Firstly, if $k$ equals to $k'$ for all points in $N_{q_l}$, then $q_{l+1}$ is simply $p^k_{i+1}$ (see Fig. \ref{fig:pca} Left). Otherwise, for all points in $ \{p^{k'}_{i'}\}_{k'\neq k\:or\:(k' = k\:and\:i' = i+1)}$, compute score $ s^{k'}_{i'}= \sigma ^k_i - v^k_i\cdot((p^{k'}_{i'}-p^k_i)/\parallel p^{k'}_{i'}-p^k_i\parallel)$, find $\hat{k},\hat{i} = \arg \mathop {\max }\limits_{k',i'} s^{k'}_{i'}$, and let $q_{l+1} = p^{\hat{k}}_{\hat{i}}$ (see Fig. \ref{fig:pca} Right).

Fig. \ref{fig:overview} (e) and (g) show the locally extracted seams, while (f) and (h) give the globally connected face curve from those local seams by the method proposed in this subsection.

\section{Experiments}
We implemented the proposed algorithm in C++, and run all the experiments on both the LFPW \cite{belhumeur2013localizing} and HELEN \cite{le2012interactive} datasets on an Intel(R) Core(TM) i7-4790 3.6GHz computer with 8GB RAM. The two face alignment benchmarks contain large variations in pose, expression, occlusion, and illumination. Each image from the LFPW dataset has 68 landmarks in total as ground-truth annotations, while the number for per image in the HELEN dataset is 194. Removing those landmarks on eyes, nose and mouth, the number of the remaining landmarks are 17 and 41 for the LFPW dataset and the HELEN dataset, respectively. We take these landmarks on face contours as the ground-truth sparse landmarks for evaluation.

To clarify, the proposed L2GSCI is not designed to handle severe occlusions (however, our method can handle small occlusions as discussed in Sec. \ref{sec:discussions}). Therefore, we filter out images with severe occlusions, and evaluate our method on the remaining 860 (out of 871) and 2290 (out of 2332) images of LFPW and HELEN. For both datasets, we randomly select 25\% images for testing and comparing, and the remaining 75\% images are for training when necessary. We summarize the above information in Table ~\ref{tab:dataset}.

\begin{table}[h]
\centering
\vspace{-0.05in}
\begin{tabular}{c|c|c|c|c}
\thickhline
Dataset & \#img & \#img-used & \#ldk & \#ldk-used \\
\thickhline
LFPW   & 871 & 860 & 68 & 17 \\
\thickhline
HELEN  & 2332 & 2290 & 194 & 41\\
\thickhline
\end{tabular}
\caption{The summarization of the two face benchmark datasets used in the paper. Here ``\#img" and ``\#img-used" indicate the number of images in the orignal dataset and those images we used in the paper, respectively. Analogously, ``\#lkd" and ``\#ldk-used" are the number of orignal sparse facial landmarks and those we used to evaluate performances, respectively.}
\label{tab:dataset}
\vspace{-0.15in}
\end{table}

For the evaluation purpose on such a new continuous face contour extraction problem in this paper, we introduce our annotation efforts and measurement metrics as follows.

\textbf{Annotation.} Besides the ground-truth sparse landmarks, we also annotate the continuous face curves for both the LFPW and HELEN datasets. For previous ground-truth sparse landmarks that may not exactly exist on our newly annotated curves, we adjust them by moving them towards to the nearest points on the curves correspondingly. Please refer to the supplemental material for our continuous and much more accurate ground truth annotations.

\textbf{Measurement.} We also introduce two measurement metrics : dense mean error (DME) with respect to ground-truth curves and sparse mean error (SME) with respect to ground-truth sparse landmarks:
\vspace{-0.075in}
\begin{itemize}
\item For DME, we compute the nearest distance of each point of an estimated curve to the corresponding ground truth curve, which are then averaged by the length of the estimated curve and normalized by the inter-ocular distance.
\vspace{-0.1in}
\item For SME, we compute the mean error between estimated and ground truth landmarks, which is also normalized by the inter-ocular distance.
\end{itemize}
\vspace{-0.075in}
To apply DME to that face alignment algorithms that produce a set of sparse landmarks, we first fit a cubic spline curve from those landmarks, and the spline is treated as the estimated continuous face contour to be compared with the ground truth curve. To apply SME to our method that produces a continuous face curve, we sample the same number of points out of the curve, and compare them with the corresponding ground truth landmarks.


\subsection{The Selection of Initial Guess}
As mentioned in Sec. \ref{sec:square-sampling}, our method relies on an initial curve that roughly indicates the position of the target face contour. Many approaches can provide us with the initialization, \eg, we can use any of the previous face alignment approaches to generate landmarks, and fit a spline curve from those landmarks. We can use the fitted spline as our initial curve.

To demonstrate more clearly about the selection process of our initial guess from the various previous face alignment approaches, we conduct the following four experiments on the HELEN dataset. For the first three experiments, we use ESR \cite{cao2014face}, LBF \cite{ren2014face}, and ERT \cite{kazemi2014one} to provide us with three kinds of initializations. Please be noted that all the three methods are trained with a full set of landmarks. We design the fourth experiment by using of ERT which however is trained with only 5 landmarks, {\em i.e.}, 1 landmark on the chin and 4 landmarks on the two cheek sides (see Fig. \ref{fig:overview}(b)).

\begin{table}[h]
\centering
\begin{tabular}{c|ccc|c}
\thickhline
Init. Methods & \tabincell{c}{ESR \\ (full)} & \tabincell{c}{LBF \\ (full)} & \tabincell{c}{ERT \\ (full)} & \tabincell{c}{ERT \\ (5 landmarks)}  \\
\hline
DME & \textbf{0.018} & 0.020 & \textbf{0.018} & \textbf{0.018} \\
SME & 0.020 & 0.022 & \textbf{0.019} & \textbf{0.019} \\
\thickhline
\end{tabular}
\caption{Initialized with four different methods. The ``full'' means the corresponding method is trained with a full set of landmarks. The ``5 landmarks'' means the corresponding method is trained with only 5 landmarks on the face contour which are shown in Fig. \ref{fig:overview} (b). The DME and SME values are statistically obtained from HELEN dataset.}
\label{tab:initialization}
\end{table}

As we can observe from Table \ref{tab:initialization}, the performances of our method with these four initializations are almost similar to each other, even though the quality of the second one is slightly a little worse than that of others. No matter using a full set of landmarks or only 5 landmarks, the performances based on ERT are the same. These observations demonstrate that our algorithm is robust against the initialization, and also suggest that ERT trained with 5 landmarks is a good choice to produce an initial guess for our algorithm.

\begin{figure*}[t]
\centering
\includegraphics[width=0.235\linewidth]{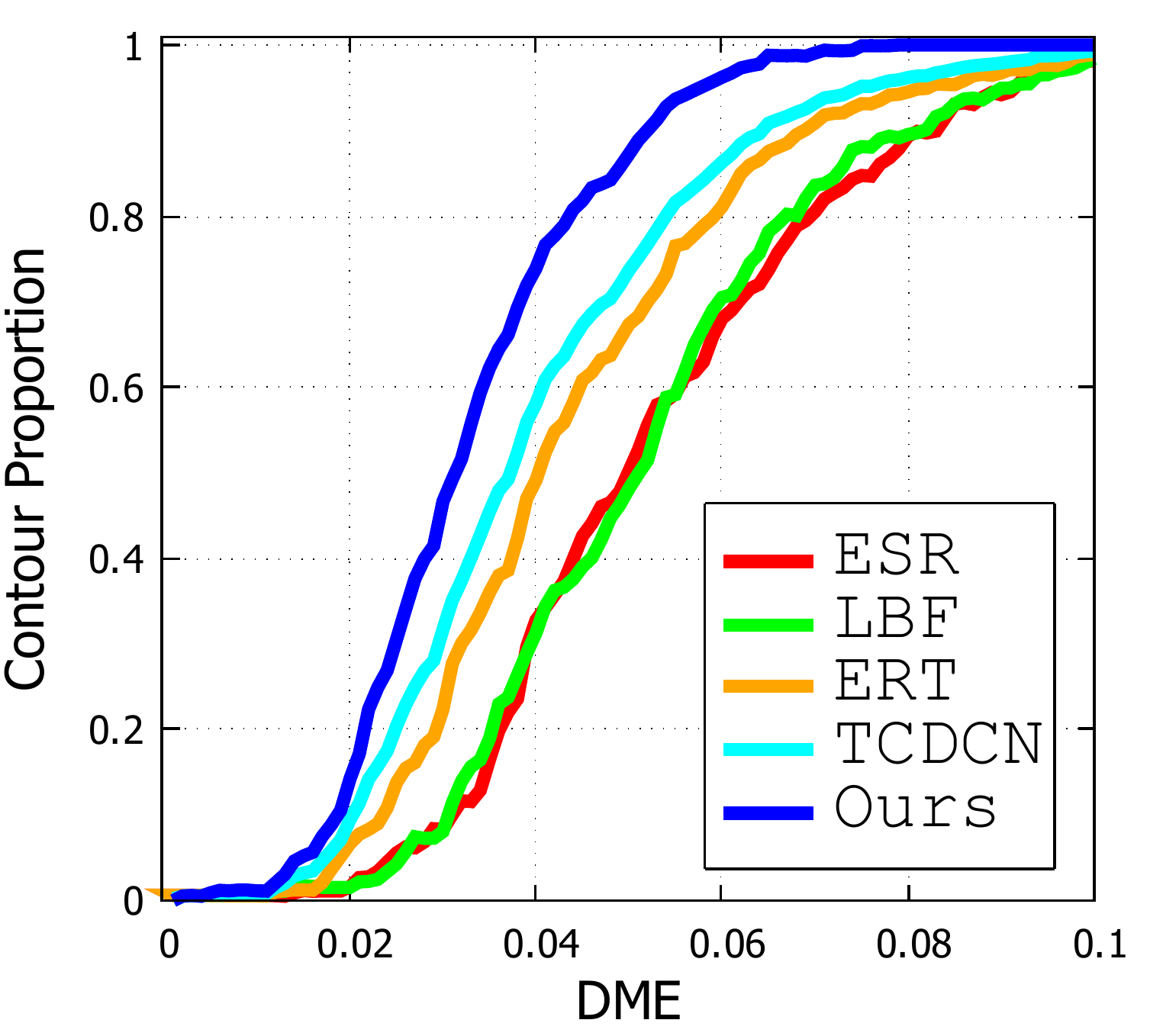}
\includegraphics[width=0.235\linewidth]{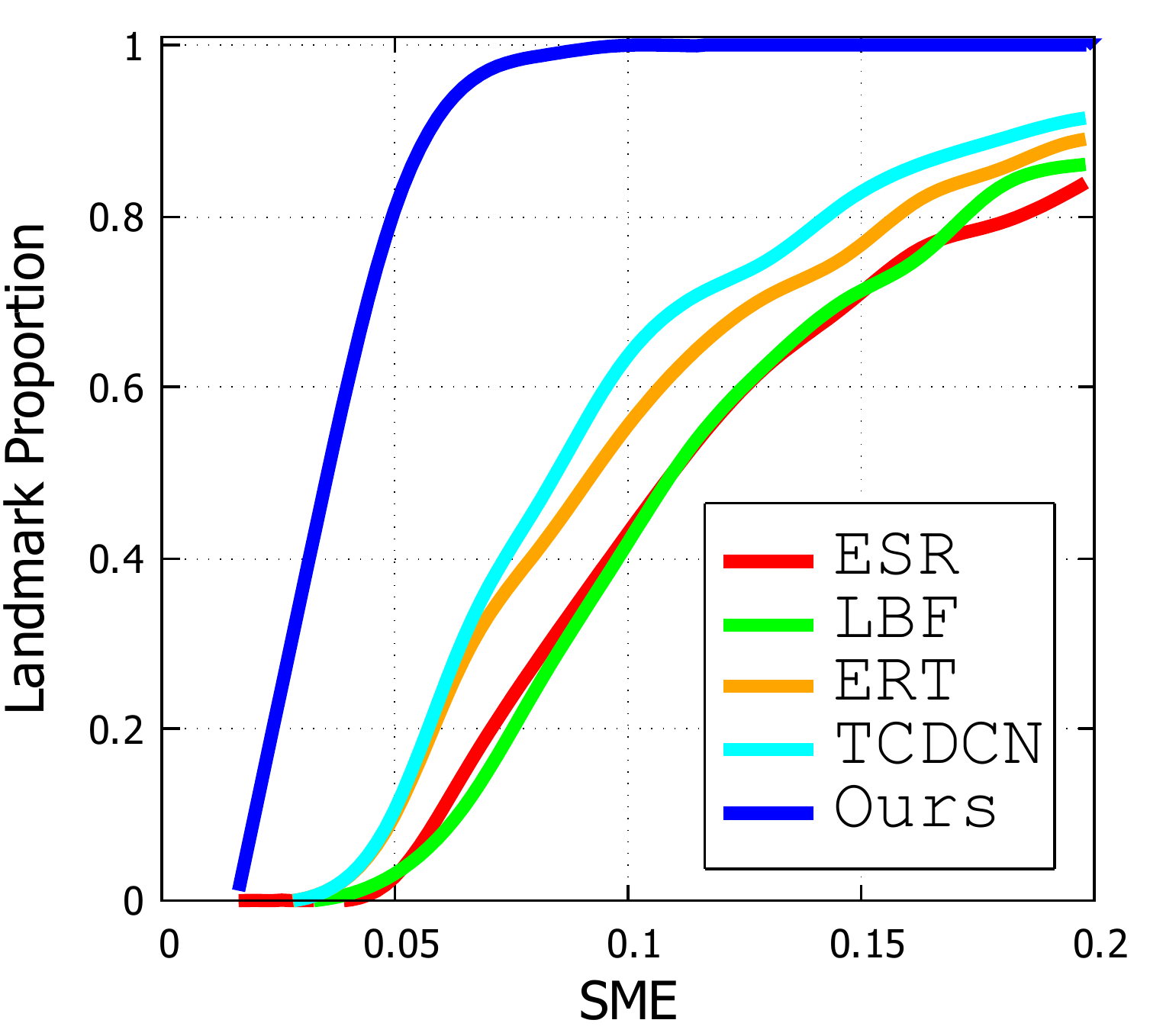}
\includegraphics[width=0.235\linewidth]{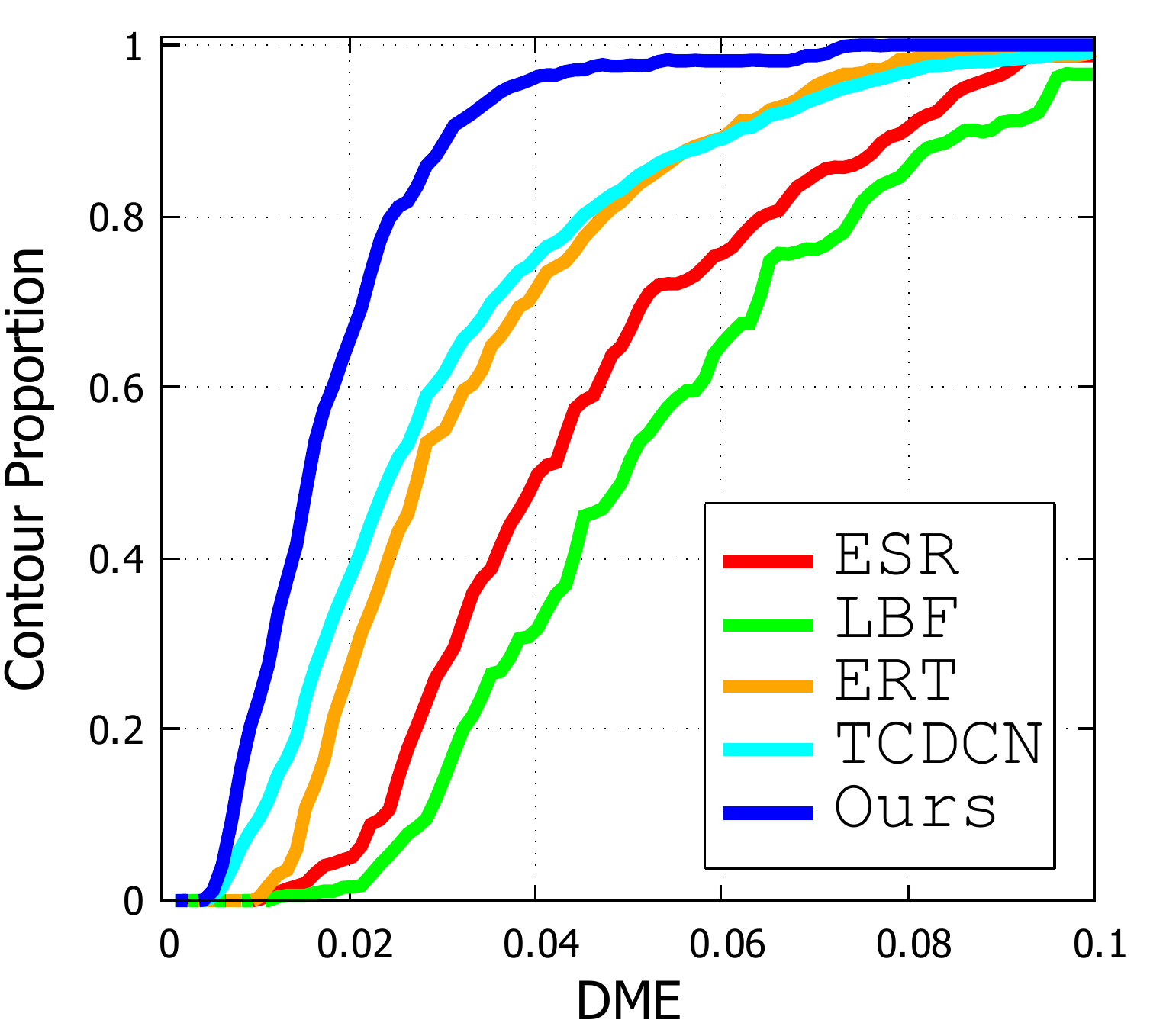}
\includegraphics[width=0.235\linewidth]{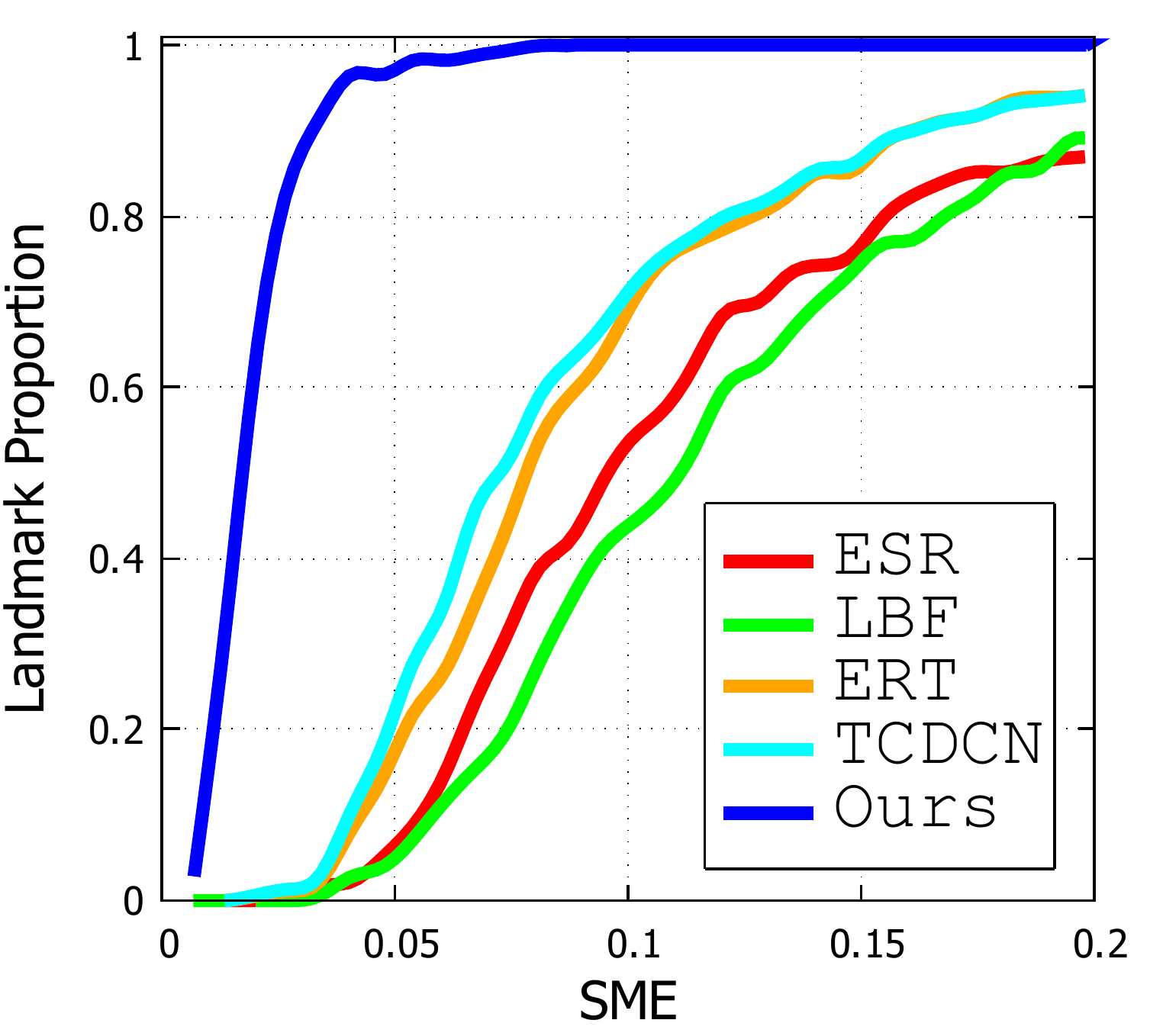}
\quad\quad (a) LFPW \hspace{2.5cm} (b) LFPW \hspace{2.5cm} (c) HELEN \hspace{2.5cm} (d) HELEN
\caption{Cumulative Error Distribution on the LFPW and HELEN datasets. Note that the higher value for contour or landmark proportion given the same DME or SME means the more accurate in performance evaluation. Our extracted face contours are more accurate than those of the four baseline approaches.}
\label{fig:CEC}
\end{figure*}

\begin{figure*}[t]
\centering
\includegraphics[width=0.12\linewidth]{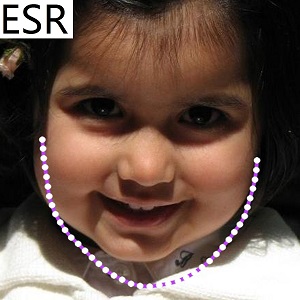}
\includegraphics[width=0.12\linewidth]{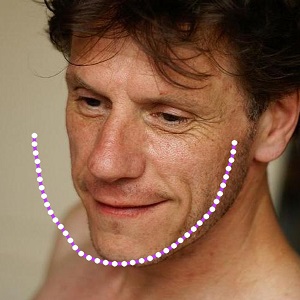}
\includegraphics[width=0.12\linewidth]{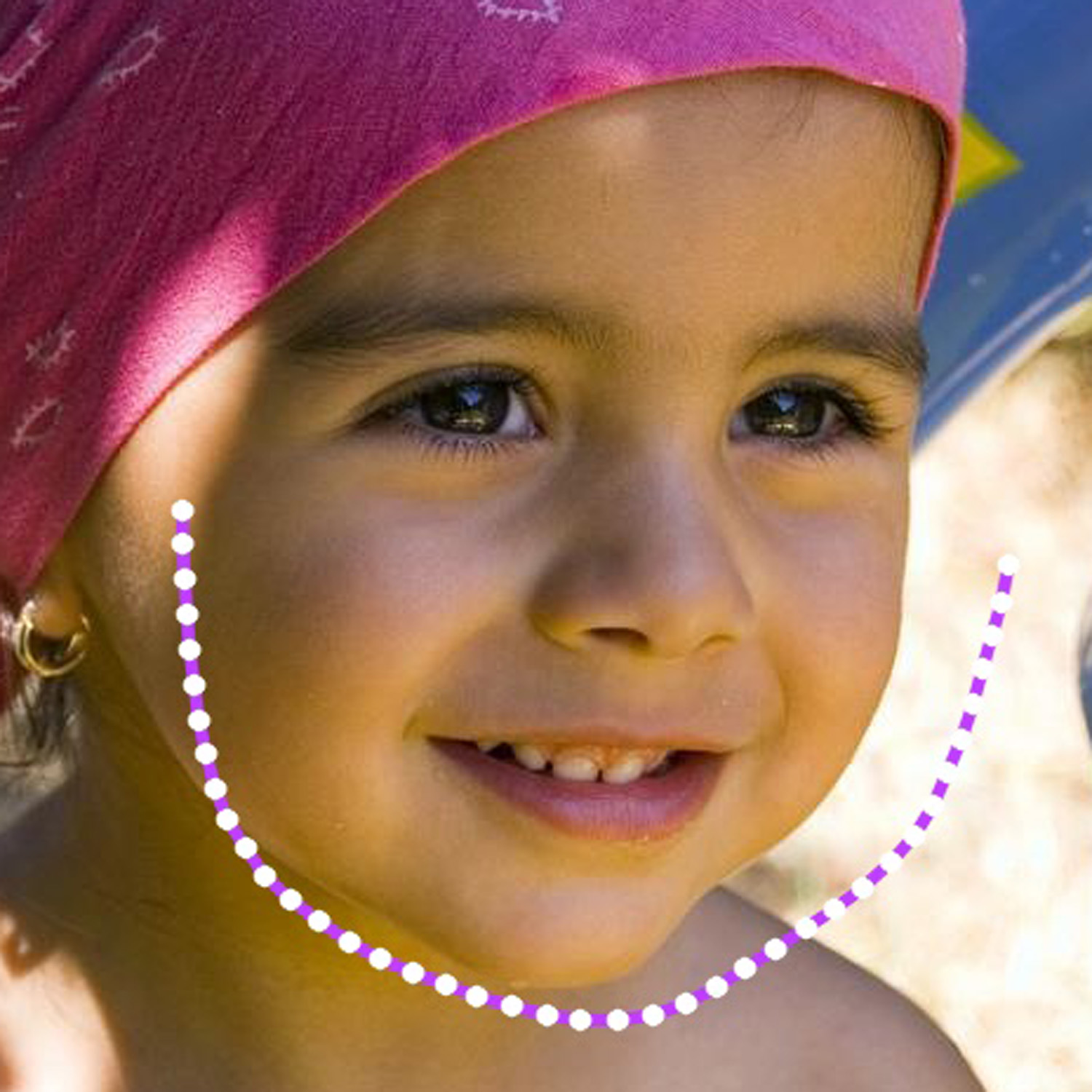}
\includegraphics[width=0.12\linewidth]{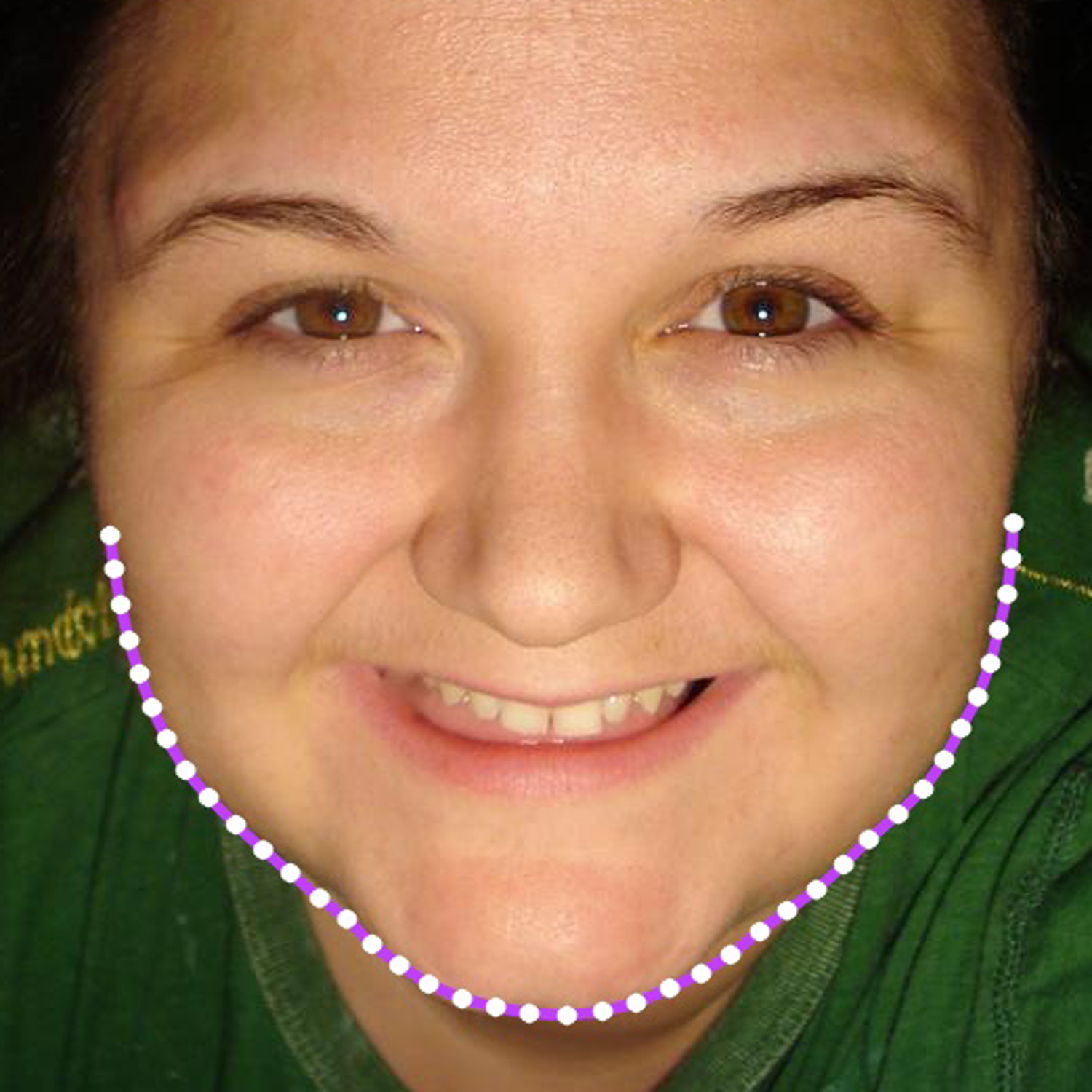}
\includegraphics[width=0.12\linewidth]{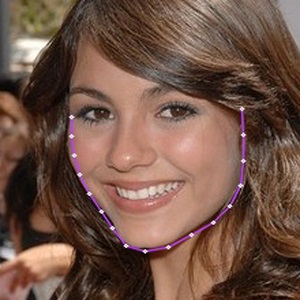}
\includegraphics[width=0.12\linewidth]{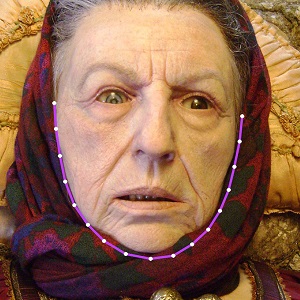}
\includegraphics[width=0.12\linewidth]{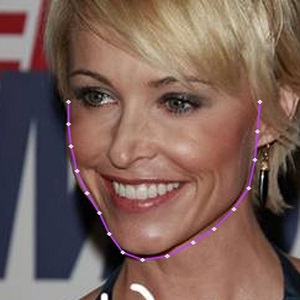}
\includegraphics[width=0.12\linewidth]{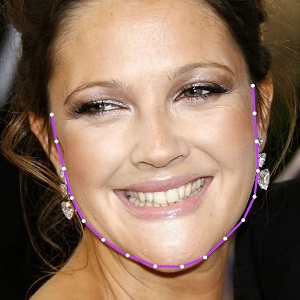}\\
\includegraphics[width=0.12\linewidth]{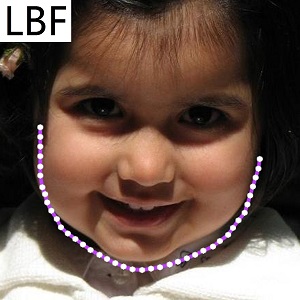}
\includegraphics[width=0.12\linewidth]{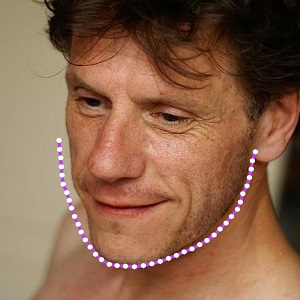}
\includegraphics[width=0.12\linewidth]{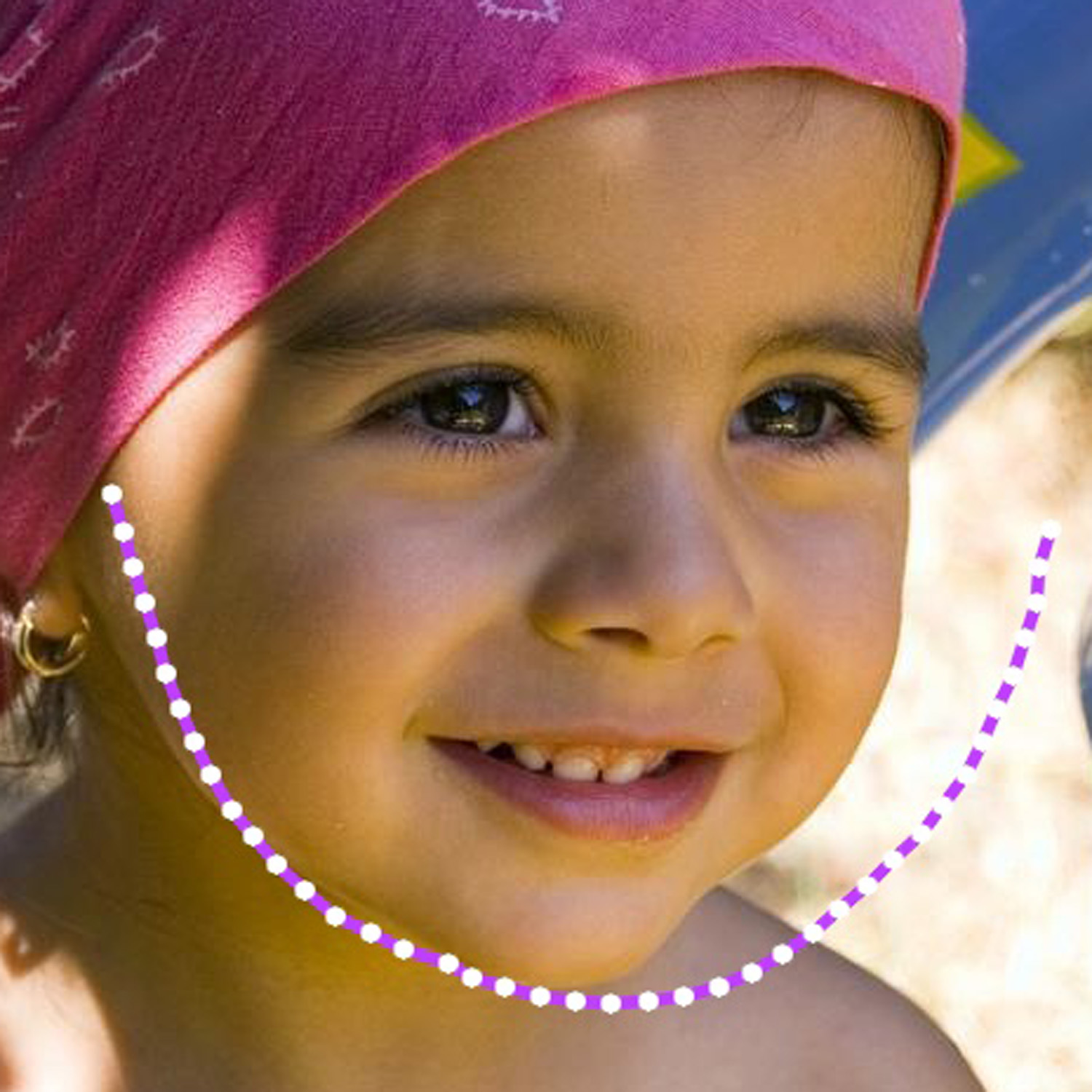}
\includegraphics[width=0.12\linewidth]{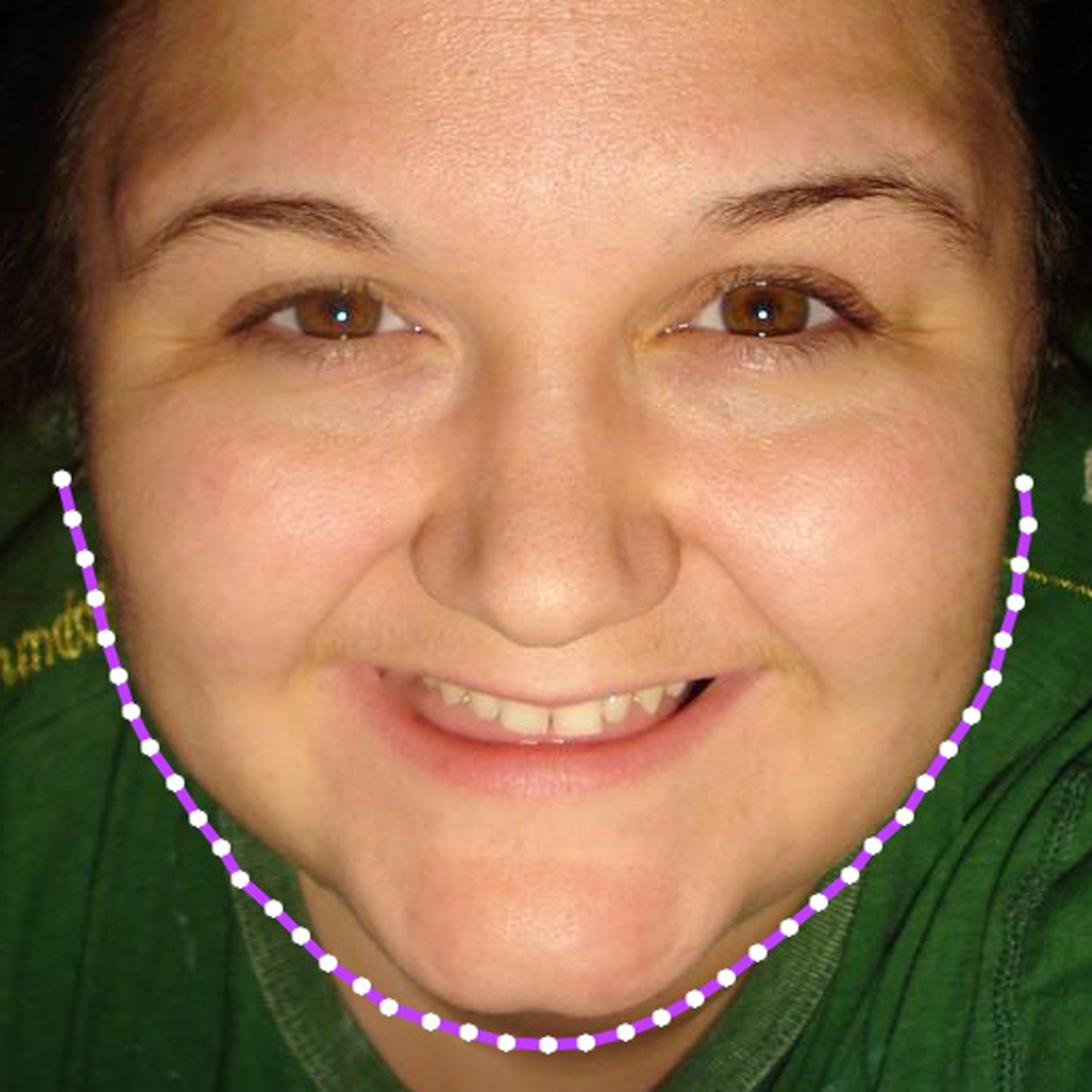}
\includegraphics[width=0.12\linewidth]{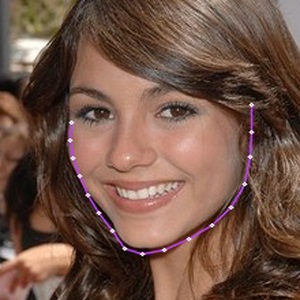}
\includegraphics[width=0.12\linewidth]{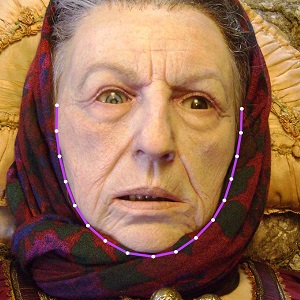}
\includegraphics[width=0.12\linewidth]{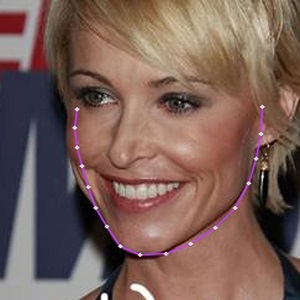}
\includegraphics[width=0.12\linewidth]{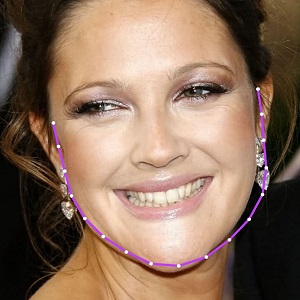}\\
\includegraphics[width=0.12\linewidth]{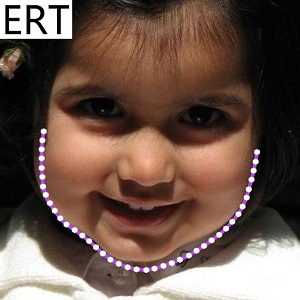}
\includegraphics[width=0.12\linewidth]{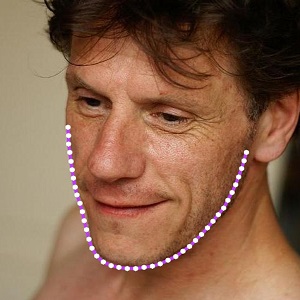}
\includegraphics[width=0.12\linewidth]{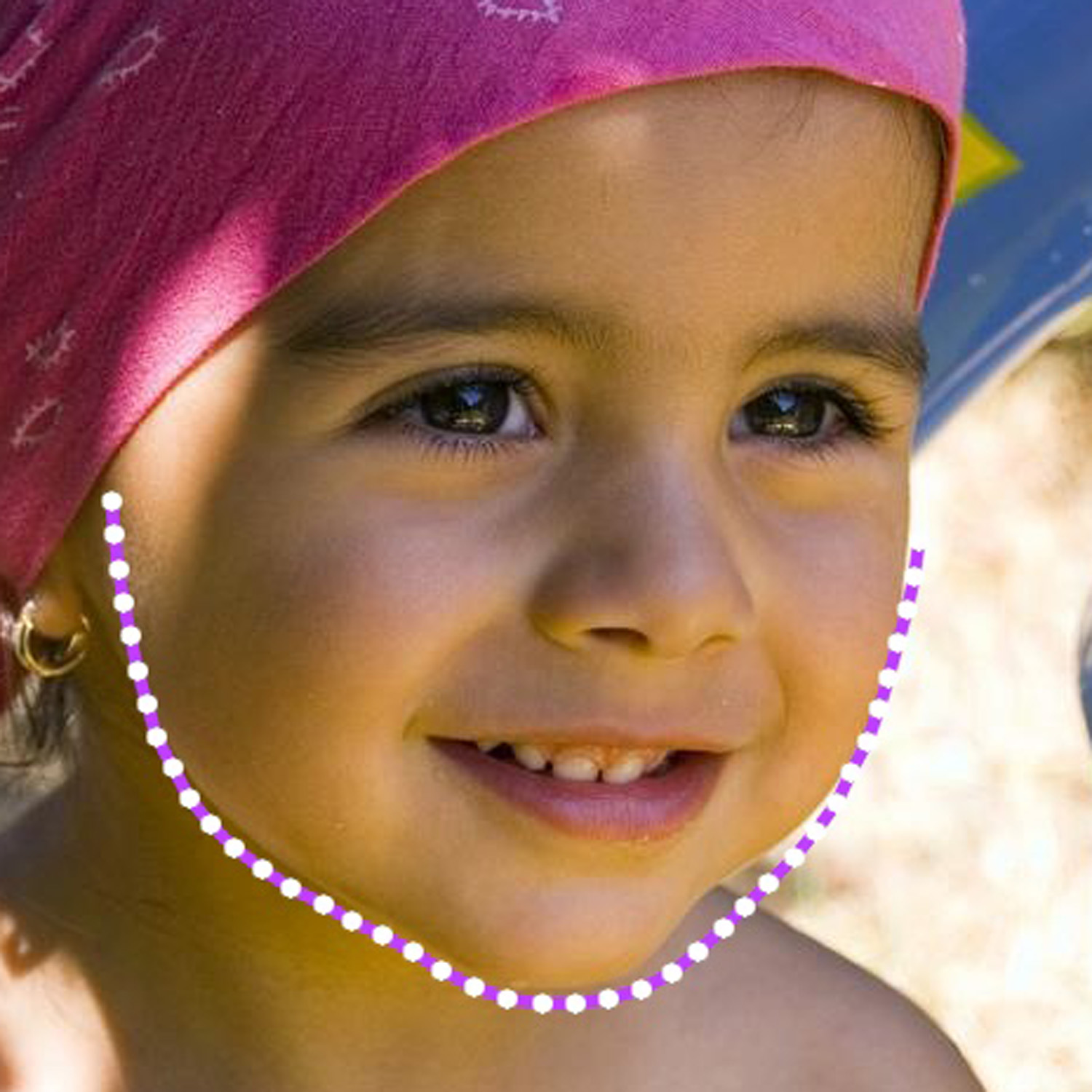}
\includegraphics[width=0.12\linewidth]{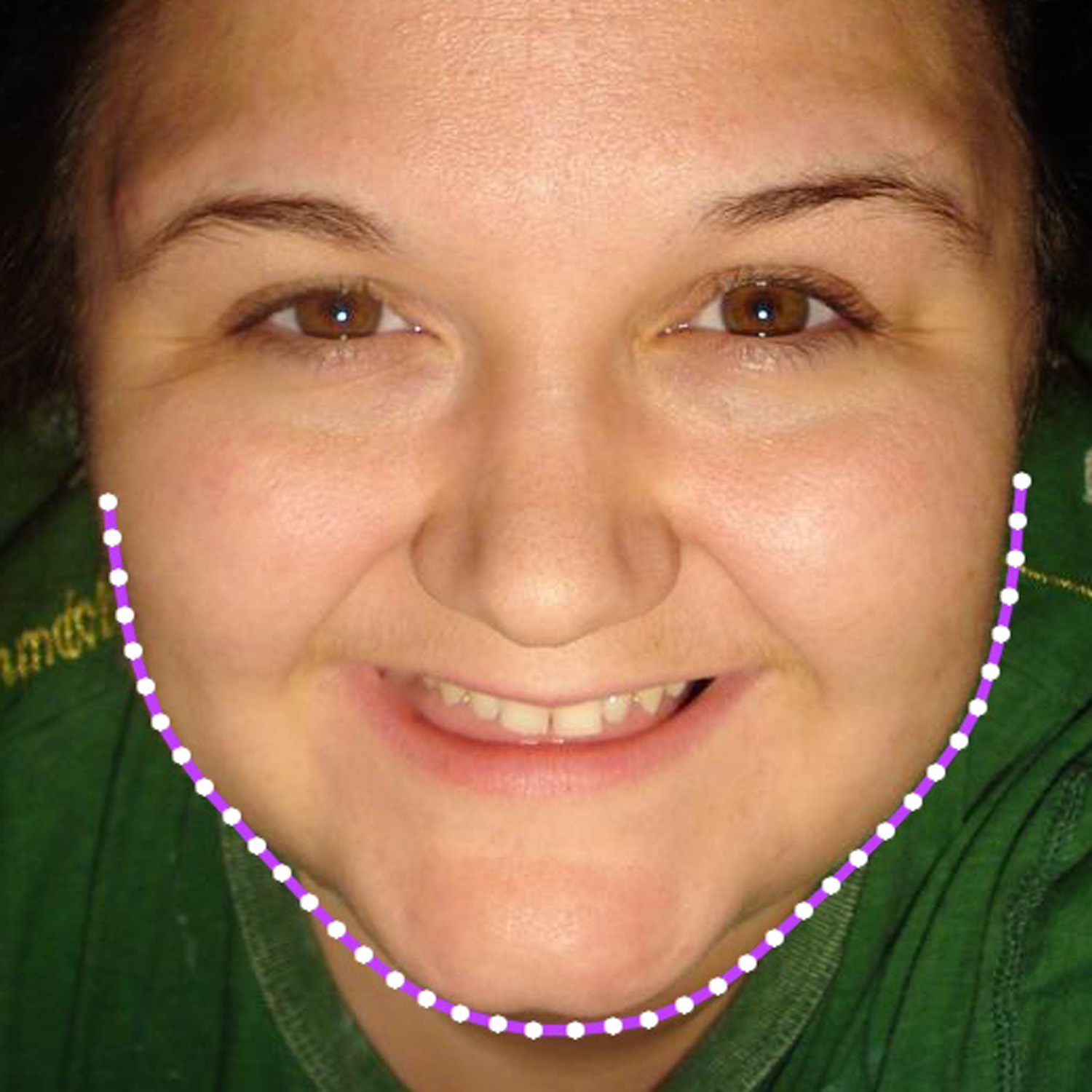}
\includegraphics[width=0.12\linewidth]{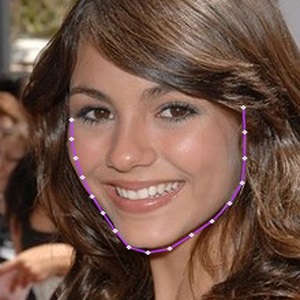}
\includegraphics[width=0.12\linewidth]{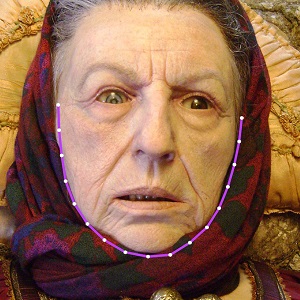}
\includegraphics[width=0.12\linewidth]{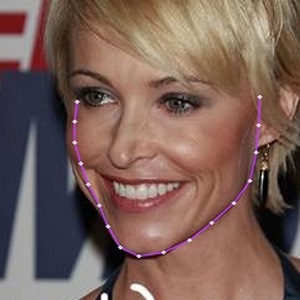}
\includegraphics[width=0.12\linewidth]{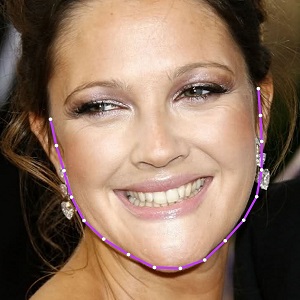}\\
\includegraphics[width=0.12\linewidth]{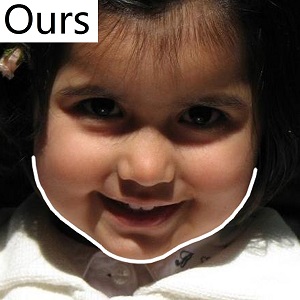}
\includegraphics[width=0.12\linewidth]{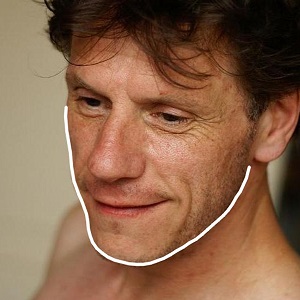}
\includegraphics[width=0.12\linewidth]{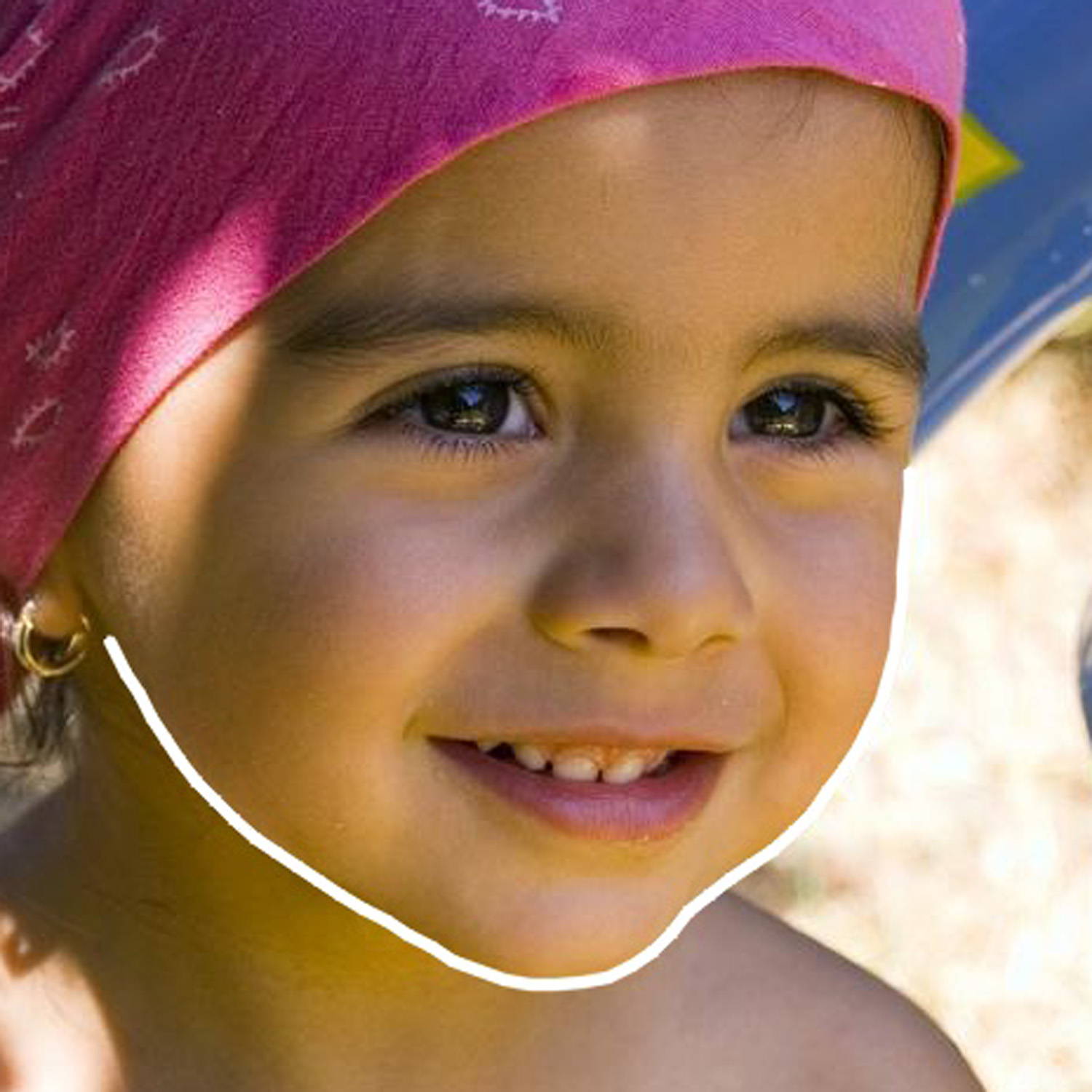}
\includegraphics[width=0.12\linewidth]{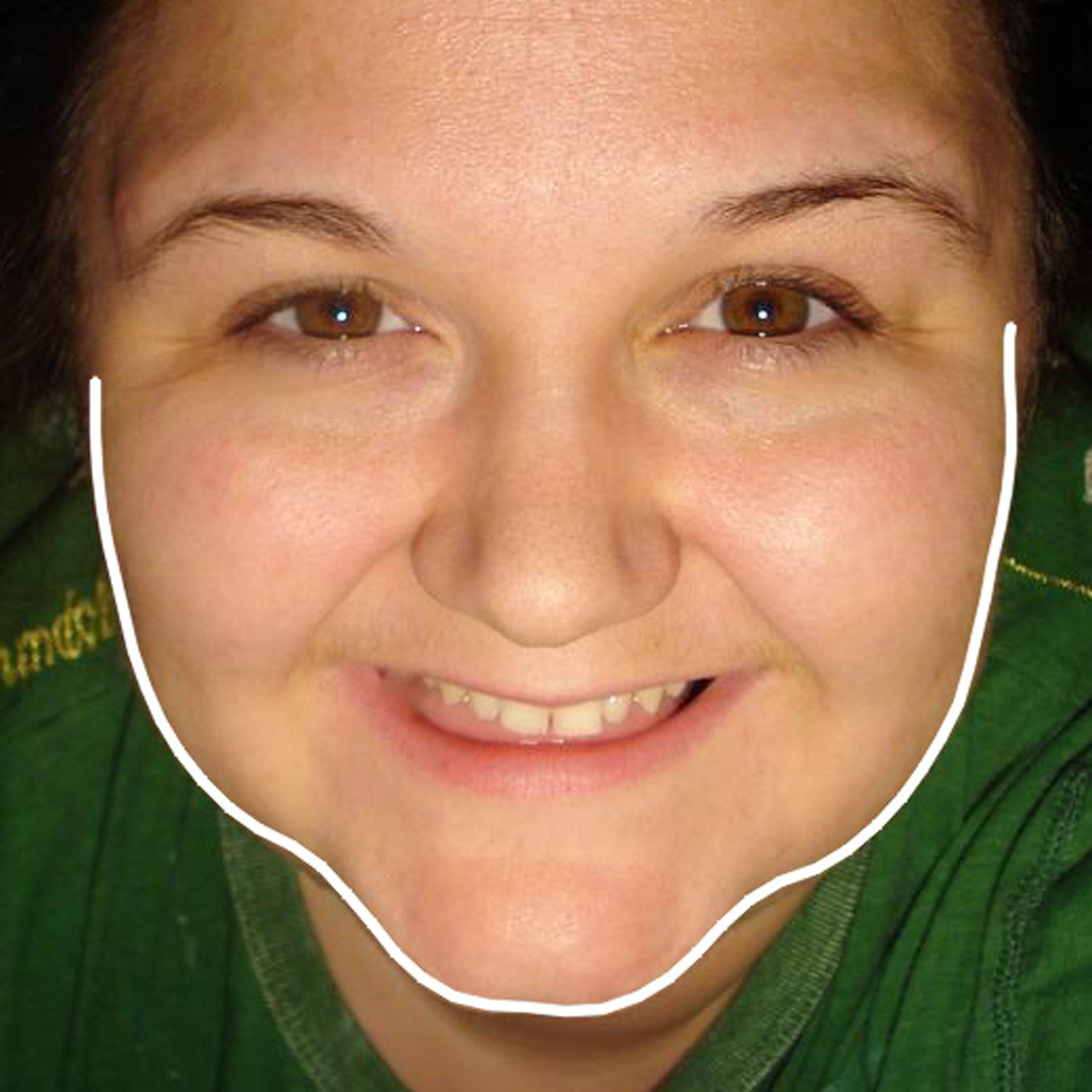}
\includegraphics[width=0.12\linewidth]{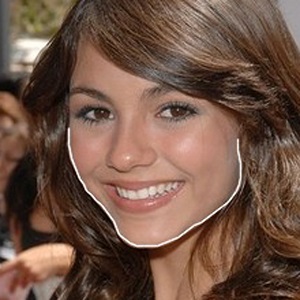}
\includegraphics[width=0.12\linewidth]{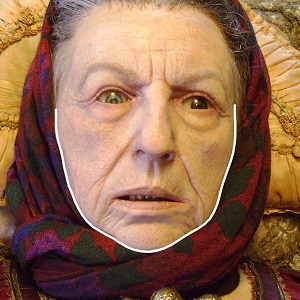}
\includegraphics[width=0.12\linewidth]{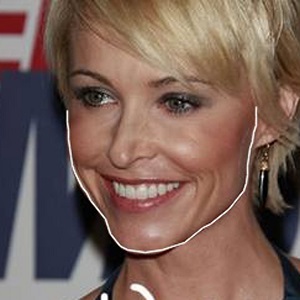}
\includegraphics[width=0.12\linewidth]{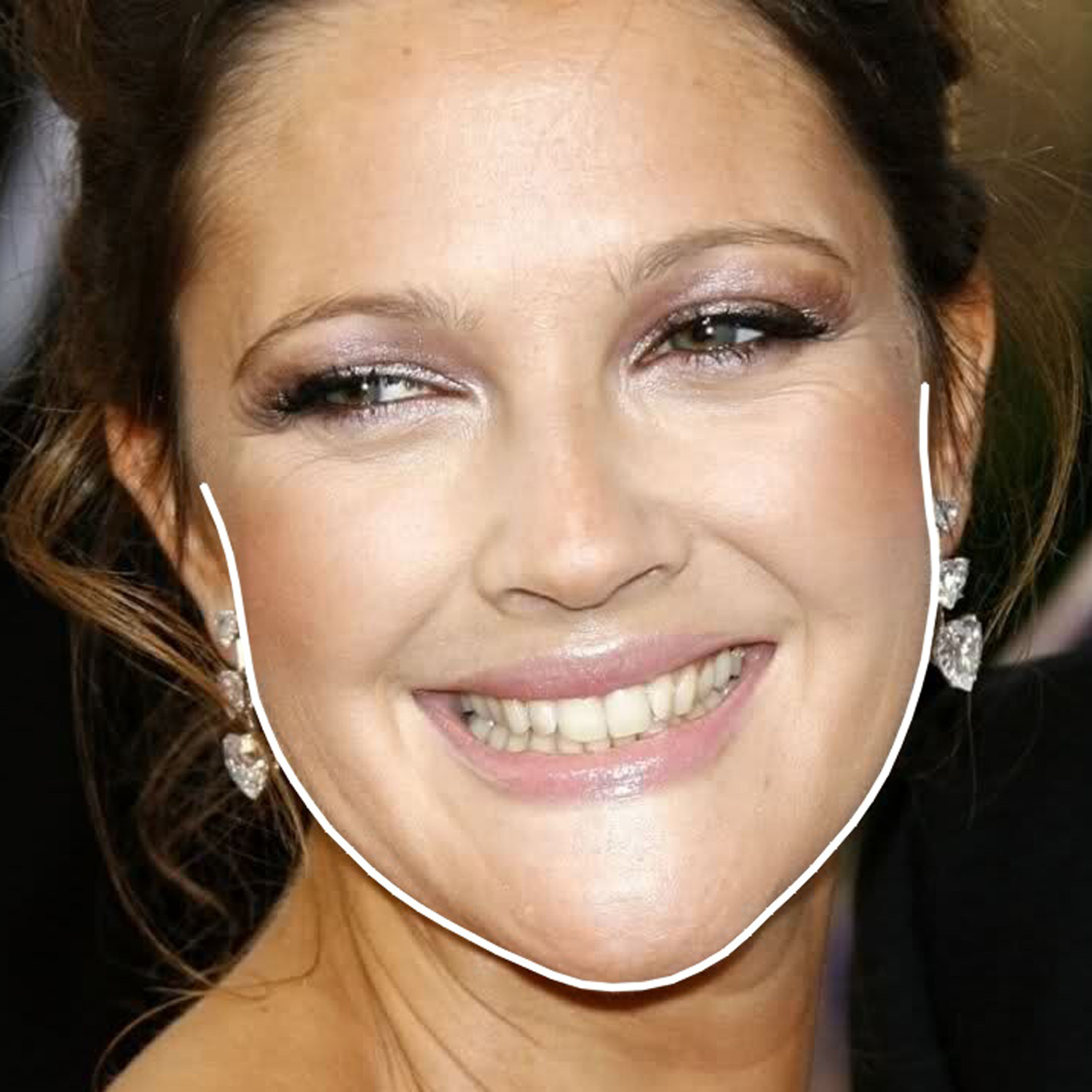}\\
\caption{Comparison between ESR \cite{cao2014face}, LBF \cite{ren2014face}, ERT \cite{kazemi2014one}, and our method. The previous methods can robustly find facial landmarks, but the landmarks are sparse. Our results are continuous curves which can reveal bending details of the real face contours. The left four columns are images from the HELEN dataset \cite{le2012interactive}, while the right four columns are images from the LFPW dataset \cite{belhumeur2013localizing}.}
\label{fig:comp}
\end{figure*}

\vspace{-0.01in}

\subsection{The Determination of Parameters}
\label{sec:determination-parameters}

Our algorithm mainly includes three parameters, {\em i.e.}, square size, square number and $\alpha$. To see how the parameters affect the performances and select the optimal value for each parameter, we conduct three experiments as follows. Due to limited space, we only report the experimental results on the HELEN dataset. Please refer to the supplemental material for the results on the LFPW dataset.

\begin{table}[h]
\centering
\begin{tabular}{c|ccccc}
\thickhline
Square Size  & 0.15x & 0.20x & 0.25x & 0.30x & 0.35x \\
\hline
DME & 0.019 & \textbf{0.018} & 0.020 & 0.020 & 0.021\\
SME & \textbf{0.019} & \textbf{0.019} & 0.021 & 0.022 & 0.023\\
\thickhline
\end{tabular}
\caption{Results with different square sizes relative to the size of the face bounding box. The DME and SME values are computed on the HELEN dataset.}
\label{tab:SquareSize}
\end{table}
\vspace{-0.1in}
\begin{table}[h]
\centering
\begin{tabular}{c|ccccc}
\thickhline
Square Num.   & 30 & 50 & 70 & 90 & 110 \\
\hline
DME & 0.019 & 0.018 & 0.018 & 0.018 & \textbf{0.017}\\
SME & 0.02  & 0.019 & 0.019 & 0.019 & \textbf{0.018}\\
\thickhline
\end{tabular}
\caption{Results with different square numbers. The DME and SME values are statistically obtained from the HELEN dataset.}
\label{tab:SquareNumber}
\end{table}
\vspace{-0.1in}
\begin{table}[h]
\centering
\begin{tabular}{c|ccccc}
\thickhline
$\alpha$   & 0.5 & 0.6 & 0.7 & 0.8 & 0.9 \\
\hline
DME & 0.019 & \textbf{0.018} & \textbf{0.018} & 0.019 & 0.020\\
SME & 0.022 & 0.022 & \textbf{0.019} & 0.021 & 0.022\\
\thickhline
\end{tabular}
\caption{Results with different $\alpha$. The DME and SME values are statistically obtained from the HELEN dataset.}
\label{tab:alpha}
\end{table}
\vspace{-0.075in}

In the first experiment, We fix the other two parameters and set five different values for the square size, {\em i.e.}, 0.15x, 0.20x, 0.25x, 0.30x and 0.35x (here `x' indicates the times multiplied to the size of the corresponding face bounding box). As we can seen in Table \ref{tab:SquareSize}, 0.20x achieves the smallest DME and SME, and both these two errors become larger with the square size increasing after 0.20x. This is due to the fact that disruptors with large gradients such as hairs and collars may come into the squares when the square size is too large, which affects the effectiveness of our method. A moderate and appropriate square size is important for high accuracy. Hence, we use 0.2x in default for the experiments in this paper.

In the second experiment, we fix the values for square size and $\alpha$, and set five square numbers, {\em i.e.}, 30, 50, 70, 90 and 110. We summarize the results on the HELEN dataset in Table \ref{tab:SquareNumber}. As we can see, 50, 70, and 90 squares achieve the same performances, which are comparable to the performance of 110 squares. We also observe that with the increasing number of squares, the efficiency reduces. Taking both the effectiveness and efficiency into consideration, we choose 50 squares as the default setting in our experiments.

In the third experiment, we set square size to be 0.20x and the square number to be 50, and run experiments on the HELEN dataset with five different values for $\alpha$, {\em i.e.}, 0.5, 0.6, 0.7, 0.8 and 0.9. Based on the observation shown in Table \ref{tab:alpha}, we set $\alpha = 0.7$ as default.

\subsection{Comparison with State-of-The-Art Methods}
To effectively and accurately evaluate the performance of our proposed method, we compare our method with three state-of-the-art regression-based approaches: ESR \cite{cao2014face}, LBF \cite{ren2014face}, and ERT \cite{kazemi2014one} on both the LFPW and HELEN datasets. We generate the results for the three approaches by running their online available codes. We also compared our method with the recently developed TCDCN \cite{zhang2016learning} which presents a deep-learning framework involving auxiliary attributes for face alignment.

\begin{table}[h]
\centering
\begin{tabular}{c|cc|cc}
\thickhline
&  \multicolumn{2}{c|}{LFPW}  & \multicolumn{2}{c}{HELEN} \\
& DME & SME & DME & SME  \\
\thickhline
ESR  & 0.052 & 0.140 & 0.045 & 0.119 \\
LBF  & 0.052 & 0.161 & 0.053 & 0.125 \\
ERT  & 0.044 & 0.124 & 0.034 & 0.097 \\
TCDCN  & 0.039 & 0.108 & 0.032 & 0.092 \\
\hline
Ours & \tabincell{c}{\textbf{0.033} \\ (15.4\%)} & \tabincell{c} {\textbf{0.037} \\ (65.7\%)} & \tabincell{c} {\textbf{0.018} \\ (43.8\%)} & \tabincell{c}{\textbf{0.019} \\ (79.3\%)} \\
\thickhline
\end{tabular}
\caption{Comparison with ESR \cite {cao2014face}, LBF \cite{ren2014face}, ERT \cite{kazemi2014one}, and TCDCN \cite{zhang2016learning} on LFPW \cite{belhumeur2013localizing} and HELEN \cite{le2012interactive} datasets. Our method has achieved 15.4\% and 65.7\% improvements upon TCDCN, measured by DME and SME on LFPW dataset respectively. The numbers become 43.8\% and 79.3\% on HELEN dataset.}
\label{tab:ComPrevious}
\end{table}

Table \ref{tab:ComPrevious} shows the comparisons between our method and the four baseline approaches. As can be seen, (1) our method outperforms the four approaches when measured by SME and DME on both datasets; (2) on the LFPW dataset, our method achieves about 15.4\% and 65.7\% improvements upon TCDCN with respective to DME and SME measurements; and on the HELEN dataset, the numbers are 43.8\% and 79.3\%; (3) SME is larger than DME due to the fact that sparse correspondences may have a certain degree of deviations, while DME that finds nearest points between two curves avoids such deviation; and (4) our approach obtains better performance on the HELEN dataset, as the images in the HELEN dataset are of higher resolution which is more favorable to our approach.

The cumulative error distributions in Fig. \ref{fig:CEC} show our advantages more apparently. This figure illustrates how much proportion of an extracted face contour (or landmarks) is within an indicated DME (or SME). We can see that our method is more superior.

Fig. \ref{fig:comp} visualizes some selected face contours extracted by our method and previous approaches. We can observe that our extracted face contours reveal the concave-convex details in the chin and cheek areas, while the results of previous methods do not contain such information. This helps explain why our proposed method is able to achieve better performance.

It worths paying attention that our proposed method takes the results of ERT as initial guess, and achieves much better performances than ERT as shown in both Table \ref{tab:ComPrevious}, Fig. \ref{fig:CEC}, and Fig. \ref{fig:comp}. These observations demonstrate that our proposed method is able to further refine the performance for the current face alignment systems.

\subsection{Runtime}
Our algorithm is very efficient. With default parameters as in Sec. \ref{sec:determination-parameters}, the average running time of our method per image is 1.1s. We also observe that most of the time is cost on computing local face segments, and thereby our algorithm can be potentially accelerated by parallel architectures.

\subsection{Discussions}
\label{sec:discussions}
Generally speaking, the existing supervised face alignment approaches can robustly find sparse landmarks, and theoretically those supervised algorithms are expected to achieve better performances given more landmark annotations and the corresponding training data. However, this is usually too expensive to fulfill. Instead of increasing the labeled training data, our method provides an economic and practical way to further refine the current performance.

\begin{figure}[t]
\centering
\includegraphics[width=0.49\linewidth]{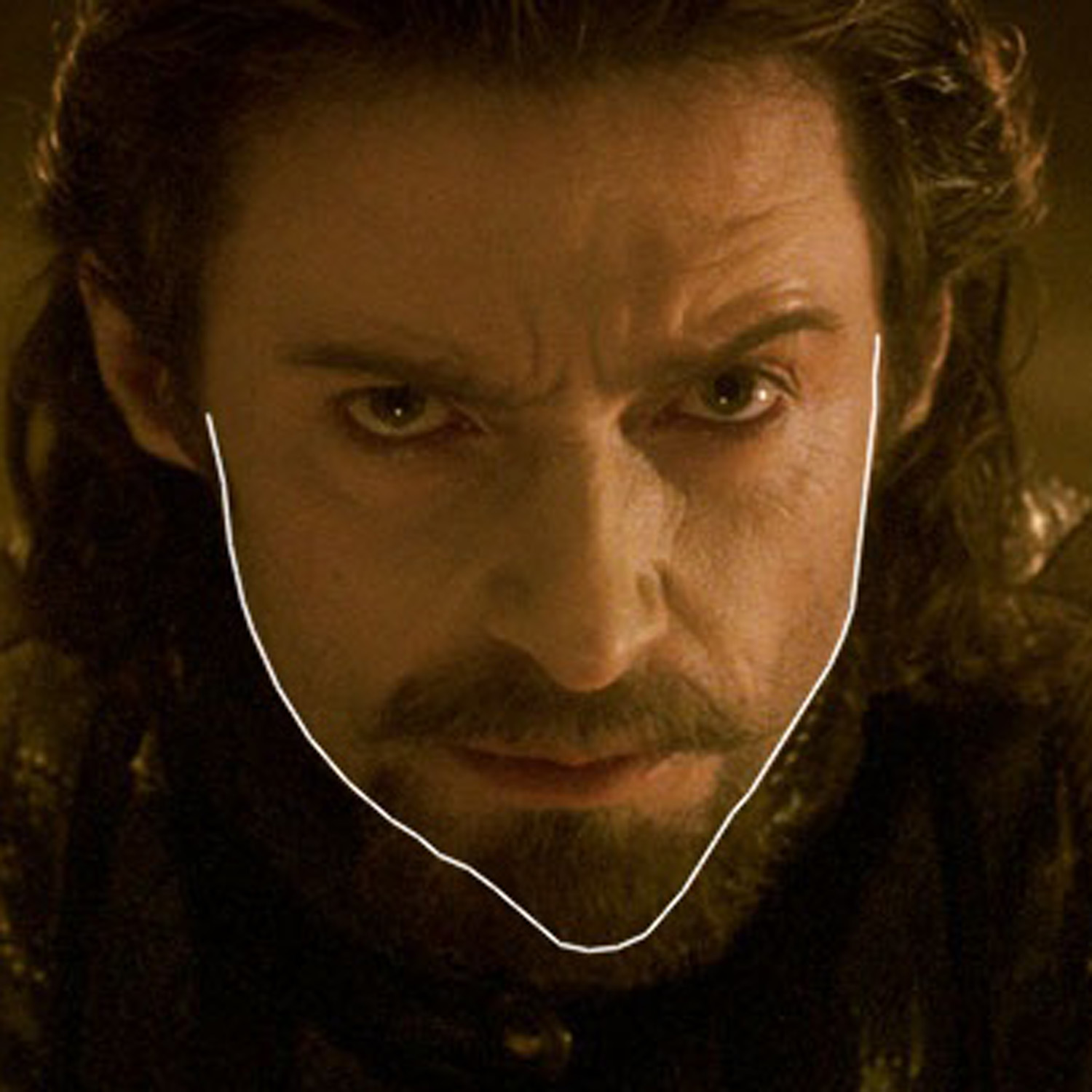}
\includegraphics[width=0.49\linewidth]{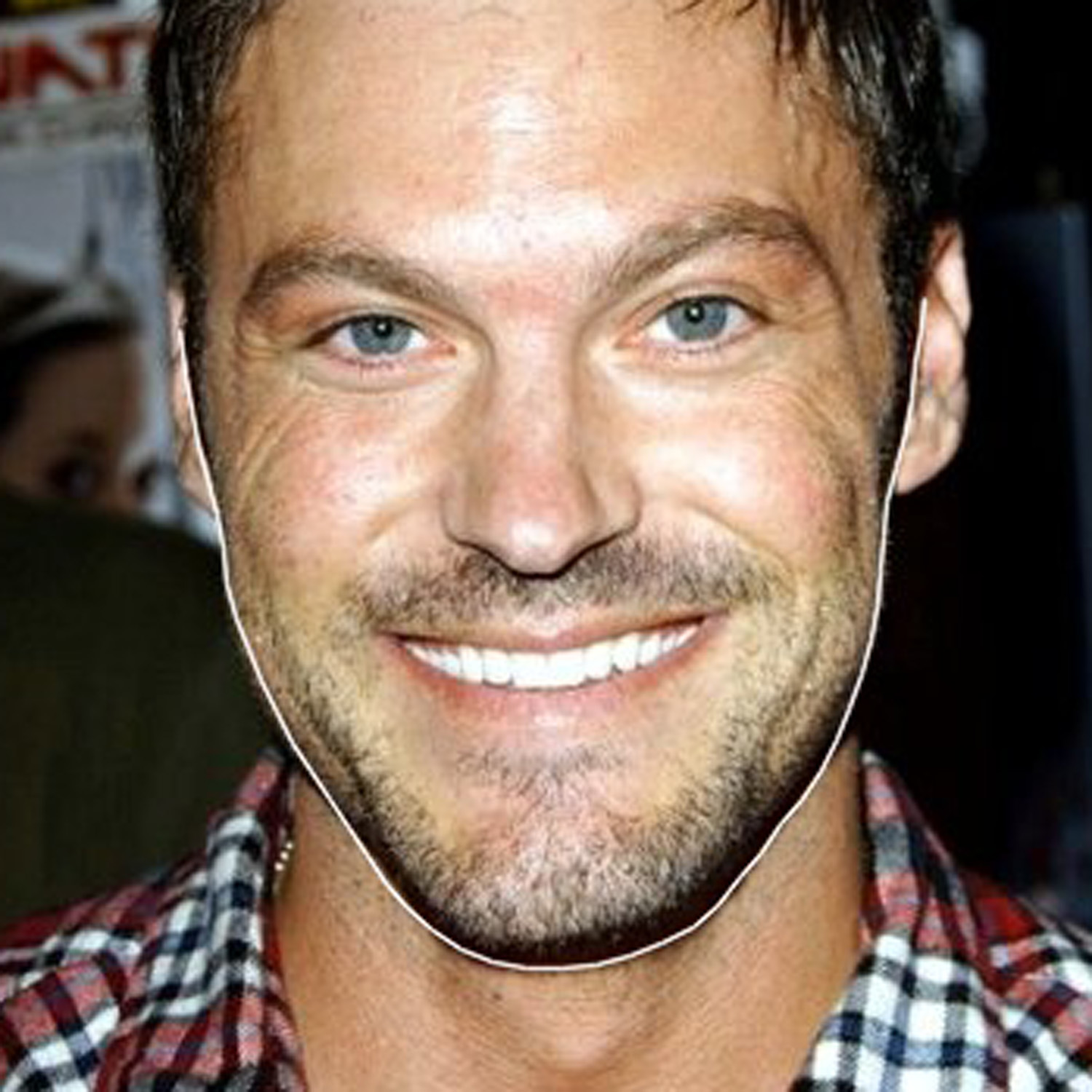}
\includegraphics[width=0.49\linewidth]{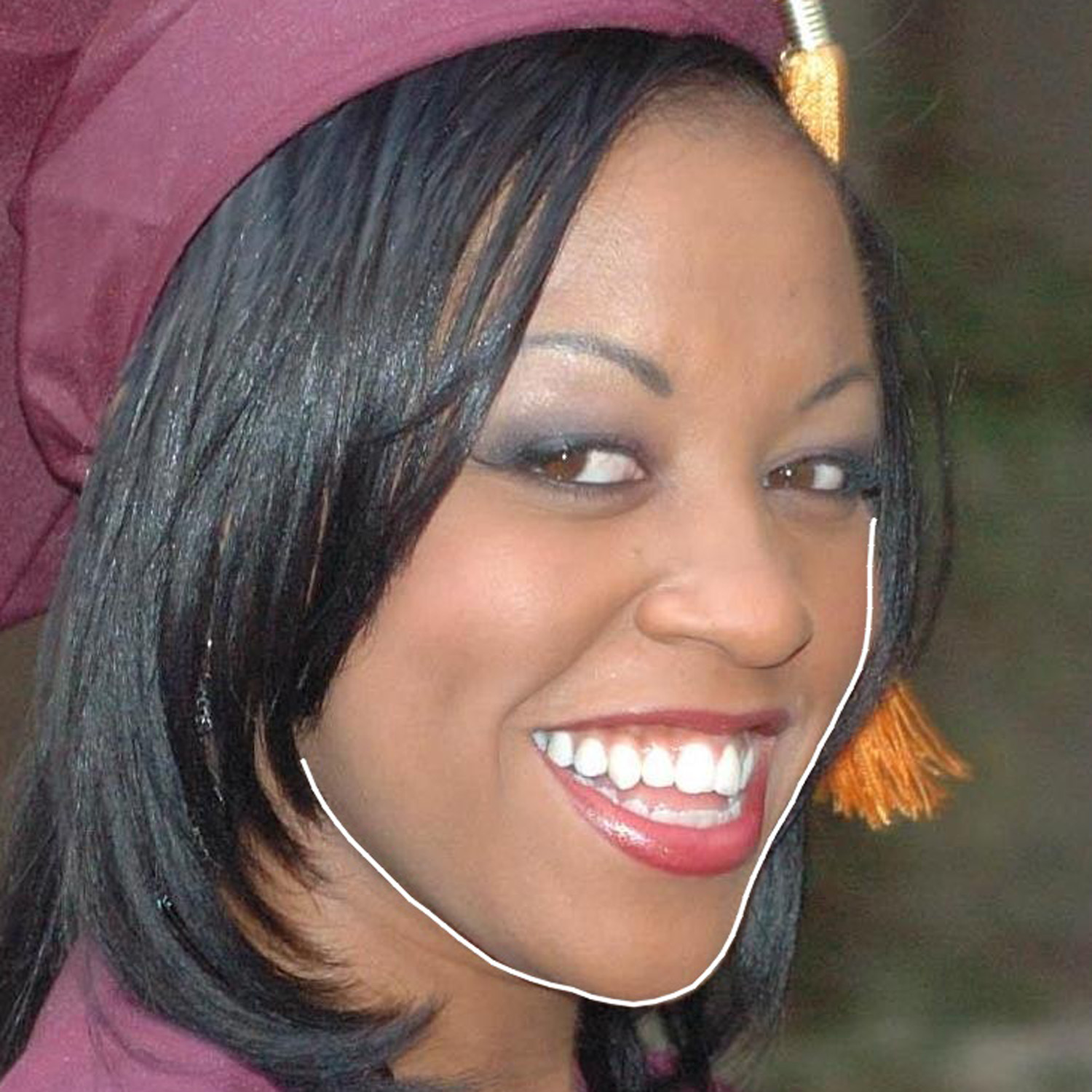}
\includegraphics[width=0.49\linewidth]{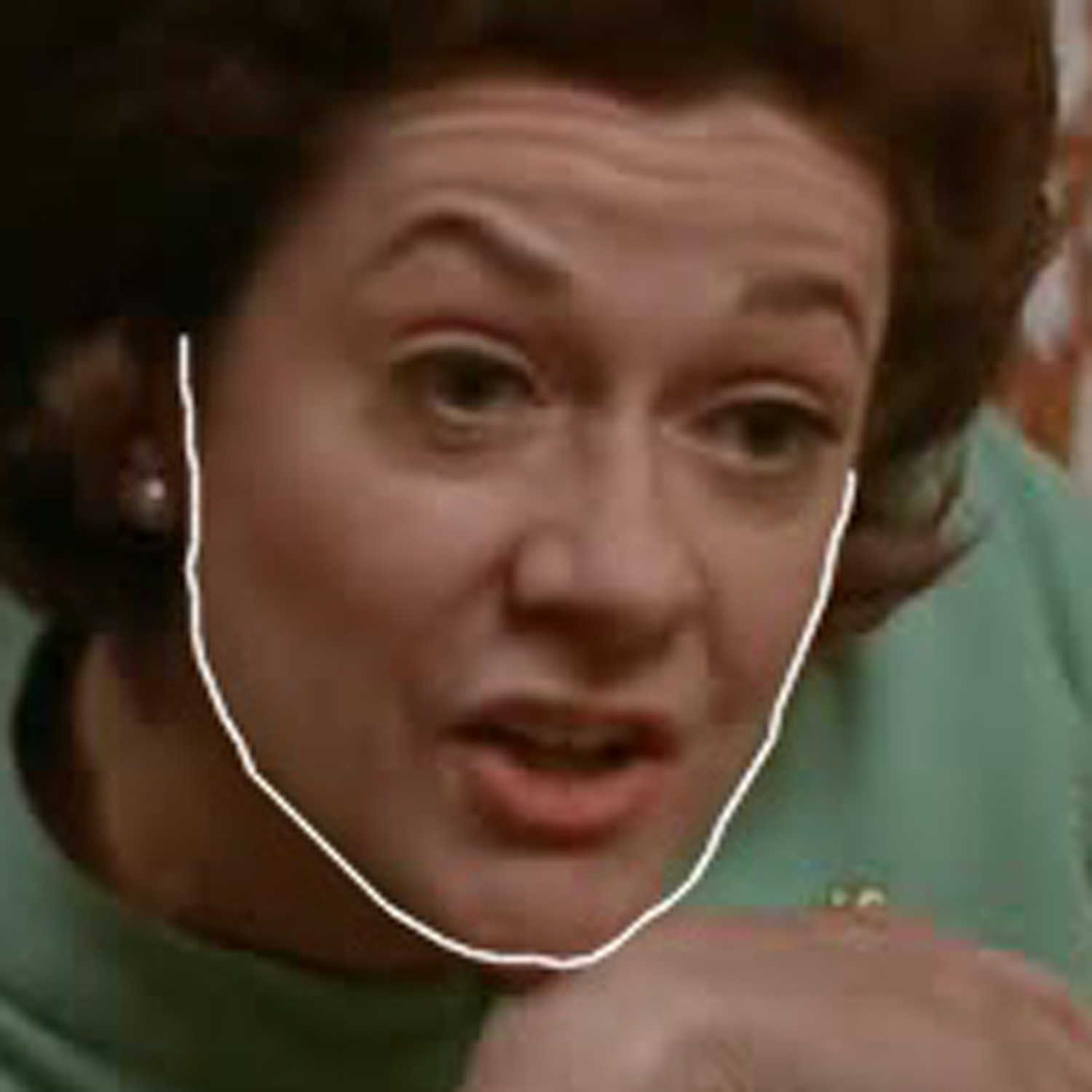}
\caption{Visualization of selected results obtained by our algorithm. (a) {\bf Top-left}: Face with big beards. (b) {\bf Top-right}: Face with small beards. (c) {\bf Bottom-left}: Face with large pose. (d) {\bf Bottom-right}: Face with dark illumination and of low resolution.}
\label{fig:Discussion}
\end{figure}

Admittedly, a limitation of our proposed method reflects in its less robustness to handle faces with the occlusion situations, especially those faces with very severe occlusions, such as very big beards or entirely occluded under hands. Fig. \ref{fig:Discussion} (a) shows a failure case. In spite of this, our method can handle small occlusions to a certain extent since the local parabola prior can provide guidance at the places where gradients are not apparent or deviate two much, as shown in Fig. \ref{fig:Discussion} (b). Except the occlusion cases, our method is able to robustly handle the other challenging cases such as various pose, expression, and illumination variations. Illustrated as two examples, the face in Fig. \ref{fig:Discussion} (c) has a large pose, and the face in Fig. \ref{fig:Discussion} (d) is dark and of low resolution.

\section{Conclusion}
We present a local-to-global parabola guided seam cutting and integrating method to extract precise and detailed face contours. As demonstrated experimentally, our approach is able to incorporate both the local and global information to robustly handle the variations of face contours. This novel method can be applied to further refine the performances of the the state-of-the-art face alignment systems, and extend their sparse facial landmarks to continuous face contours. Our future work is to extend our current work to address challenging vision problems like face recognition, object and edge detection.

{\small
\bibliographystyle{ieee}
\bibliography{egbib}
}

\end{document}